\documentclass[journal,twoside]{IEEEtran}

\usepackage[numbers,sort&compress]{natbib}
\usepackage[pdftex]{graphicx}
\usepackage{color}
\usepackage{xcolor}
\usepackage{hyperref}
\definecolor{RED}{rgb}{1,0,0}
\definecolor{darkred}{rgb}{0.6,0,0}
\definecolor{darkgreen}{rgb}{0,0.6,0}
\definecolor{darkblue}{rgb}{0,0,0.6}
\hypersetup{hidelinks, breaklinks, colorlinks,
  linkcolor={darkred}, citecolor={darkgreen}, urlcolor={darkblue}}
\usepackage{url}
\usepackage{amsmath}
\usepackage{amssymb}
\usepackage{paralist}

\usepackage{subcaption}

\graphicspath{{images/}}
\DeclareGraphicsExtensions{.png,.jpg}

\newcommand{\eg}{e.g.,}
\newcommand{\ie}{i.e.,}

\newcommand{\etc}{etc.}
\newcommand{\TM}{\textsuperscript{\scriptsize TM}}
\newcommand{\Figref}[1]{Figure~\ref{#1}}
\newcommand{\figref}[1]{Figure~\ref{#1}}
\newcommand{\Tabref}[1]{Table~\ref{#1}}
\newcommand{\tabref}[1]{Table~\ref{#1}}

\newcommand{\secref}[1]{Section~\ref{#1}}
\newcommand{\degree}{$^{\circ}$}
\newcommand{\ignore}[1]{}
\newcommand{\needswork}[1]{\textbf{\textcolor{red}{#1}}}

\newcommand{\equref}[1]{Equation~\ref{#1}}

\newcommand{\wander}{\textsc{Wander}}
\newcommand{\approachperson}{\textsc{Approach\_person}}
\newcommand{\holdconversation}{\textsc{Hold\_conversation}}
\newcommand{\followdirections}{\textsc{Follow\_directions}}
\newcommand{\navigatedoor}{\textsc{Navigate\_door}}

\newcommand{\makedecision}{\textsc{Make\_decision}}
\newcommand{\rotaterecovery}{\textsc{Rotate\_recovery}}
\newcommand{\rotate}{\textsc{Rotate}}
\newcommand{\driveforward}{\textsc{Drive\_forward}}
\newcommand{\drivethroughintersection}{\textsc{Drive\_through\_intersection}}
\newcommand{\complete}{\textsc{Complete}}

\newsavebox{\spaceofwidthbox}
\newcommand{\spaceofwidth}[1]%
{\savebox{\spaceofwidthbox}{#1}\hspace*{\wd\spaceofwidthbox}}
\newcommand{\digit}{\spaceofwidth{0}}

\setdefaultenum{a)}{}{}{}

\begin{document}

\title{The Amazing Race\TM: Robot Edition}

\author{Jared~Sigurd~Johansen,~\IEEEmembership{Graduate Student Member,~IEEE,}
  Thomas~Victor~Ilyevsky,~\IEEEmembership{Graduate Student Member,~IEEE,}
  and~Jeffrey~Mark~Siskind,~\IEEEmembership{Senior Member,~IEEE}% <-this % stops a space
  \thanks{Manuscript received \needswork{MM dd, YYYY}; revised
    \needswork{MM dd, YYYY}.
    This work was supported, in part, by the US National Science Foundation
    under Grants 1522954-IIS and 1734938-IIS, and by the Intelligence
    Advanced Research Projects Activity (IARPA) via Department of
    Interior/Interior Business Center (DOI/IBC) contract number D17PC00341.
    \emph{(Jared Sigurd Johansen and Thomas Victor Ilyevsky contributed
      equally to this work.)
      (Corresponding author: Jared Sigurd Johansen.)}}% <-this % stops a space
  \thanks{All authors were with the School of Electrical and Computer
    Engineering, 465 Northwestern Avenue, Purdue University, West Lafayette,
    IN, 47907 USA e-mail:
    \texttt{\{\href{mailto:jjohanse@purdue.edu}{jjohanse},\href{mailto:tilyevsk@purdue.edu}{tilyevsk},\href{mailto:qobi@purdue.edu}{qobi}\}@purdue.edu}.
  }% <-this % stops a space}
}

\markboth{IEEE Transactions on Robotics,~Vol.~\needswork{VV}, No.~\needswork{NN}, \needswork{MM}~\needswork{YYYY}}%
{The Amazing Race\TM: Robot Edition}

\maketitle

\begin{abstract}
  State-of-the-art natural-language-driven autonomous-navigation systems
  generally lack the ability to operate in real unknown environments without
  crutches, such as having a map of the environment in advance or requiring a
  strict syntactic structure for natural-language commands.
  Practical artificial-intelligent systems should not have to depend on such
  prior knowledge.
  To encourage effort towards this goal, we propose The Amazing Race\TM:
  Robot Edition, a new task of finding a room in an unknown and unmodified
  office environment by following instructions obtained in spoken dialog from
  an untrained person.
  We present a solution that treats this challenge as a series of sub-tasks:
  natural-language interpretation, autonomous navigation, and semantic mapping.
  The solution consists of a finite-state-machine system design whose states
  solve these sub-tasks to complete The Amazing Race\TM.
  Our design is deployed on a real robot and its performance is demonstrated
  in 52 trials on 4 floors of each of 3 different previously unseen buildings
  with 13 untrained volunteers.
\end{abstract}


\IEEEpeerreviewmaketitle

\section{Introduction}

\IEEEPARstart{T}{he} Amazing Race\TM\ is a popular reality television show in
which two-person teams race to some designated location.
They typically have to figure out where they are, navigate through foreign
areas, and ask people for directions to their destination.
State-of-the-art artificial-intelligence and human-robot-interaction research
enables robots to solve such natural-language-driven navigation tasks
\citep{hanheide2017robot, bauer2009autonomous, matuszek2010following,
  oh2015toward, kollar2010toward, hemachandra2015learning, oh2015toward,
  thomason2015learning, barrett2018driving}.
However, these systems suffer from certain limitations such as requiring a
specific syntactic structure for natural-language commands or requiring a map
of the environment.
A system design without these crutches is crucial for operating autonomously in
new, unknown environments.
Additionally, it is important for these systems to have productive interaction
with humans for the purpose of receiving instructions to execute or to learn
new information.
However, it is impractical and inconvenient for every person to know the
precise details of a robotic system in order to interact with it.
Therefore, robotic system designs should be based on and thereby exhibit
human-like behavior to enable natural and useful interaction with humans.

To drive research towards this outcome, we propose a novel challenge problem to
the AI community: The Amazing Race\TM.
The task is this: we place the robot in an unknown environment, without a map,
and give it the name of a person, room, or building to find.
For the purpose of constructing an initial solution to this task, we restrict
the goal to finding a door with a specified number on a single floor of a
building.
Because the robot has no prior knowledge about the goal location nor the
structure of the environment, it must seek a person for assistance.
Once a person is found, the robot must engage them in a dialogue to obtain
directions to the goal.
It has to follow these directions to reach the hallway that the room is in, at
which point it would have to systematically search for doors and read their
door tags to locate the correct one.
We present a novel finite-state-machine (FSM) system design that
makes these logical steps to accomplish this task.

Our hypothesis is that this problem requires a specific set of abilities
that are invoked in a particular order, equivalent to an FSM\@.
We carefully chose a sequence of specific states for our FSM design that
reflects the steps a typical person would take to efficiently find a room in an
unknown office environment.
Initially knowing nothing about the environment, many people would ask the
first person they see for directions.
After receiving these directions and potentially asking for clarification, they
would follow them to the approximate location of the room and begin looking for
the specific number.
Our FSM design mimics this human behavior, which can result in a shorter and
more efficient path to the goal as opposed to a simple exhaustive search of the
environment.
Although an exhaustive search would eventually succeed, it would not
necessarily take the most efficient path, would ignore people who may have a
rich understanding of the environment and a willingness to help, and would not
push the envelope on robot cognition.
Additionally, an exhaustive search becomes unsuitable as the size of the
environment increases.

\begin{figure*}
  \centering
  \includegraphics[width=\linewidth]{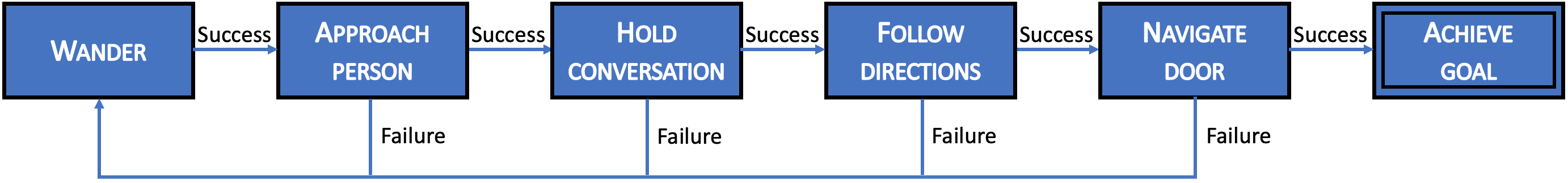}
  \caption{Finite-state-machine view of the system architecture.}
  \label{fig:brain-fsm}
\end{figure*}

In order to enable such human-like reasoning, we abstract low-level sensor data
such as audio, video, and LiDAR into information that a human would have
readily available such as navigation directions and potential goal locations.
The states in our FSM design, as shown in \figref{fig:brain-fsm}, also referred
to as \emph{behaviors}, make use of this information to handle new and/or
complex situations.
They also incorporate methods to handle complete, partial, or erroneous
information similar to how a human would when encountering such in
conversation or navigation.
We believe this design mimics high-level aspects of human behavior on the same
task.
Additionally, to the best of our knowledge, this paper makes the following novel
contributions:
\begin{compactitem}
\item It is the first to propose a method for constructing a navigation plan by
  extracting directions and intersections from spoken natural-language dialogue
  with a person.
  This navigation plan is executed by the robot using novel autonomous methods
  that ground directions and intersections in the map constructed by a SLAM
  algorithm.
\item It is the first to propose a method for detecting, localizing, and
  inspecting doors to find the desired goal.
  It employs novel common-sense reasoning to efficiently search the environment
  for the correct door.
\end{compactitem}

In order to evaluate the performance of our system, we conducted 52 trials
across 4 floors of each of 3 buildings previously unseen by the robot.
We recruited 13 untrained volunteers to provide the robot with directions to
the specified goal room number in each trial.
Due to our system's ability to recover from individual behavior failures, we
demonstrate a high success rate of 76.9\%.
Additionally, we document observations made during the trials as opportunities
for further research into this task.

\section{System Overview}

In this section, we provide a high-level overview of our system's hardware,
software, and architecture as well as the method used to construct a
qualitative map from the quantitative map produced by SLAM\@.
We provide detailed descriptions of our system's components in the next
section.

\subsection{Hardware and Software}

Our robotic platform consists of a Clearpath Husky A200\TM\ UGV equipped with
an Open IMU UM7, Velodyne VLP-16 3D LiDAR, Axis M5525-E PTZ Camera, Blue Yeti
microphone, and a System76 Laptop with two Nvidia GeForce GTX 1080 GPUs.
Each of these is a commercial, off-the-shelf product.
Clearpath integrated the IMU, LiDAR, PTZ Camera, and laptop onto the Husky
A200\TM\ UGV\@.

Our software\footnote{All software used to produce the results in this
  manuscript is available at \url{https://github.com/qobi/amazing-race}.} is
implemented in a combination of C++ and Python, using ROS Kinetic as the
communication framework between different components of the system.
We use Google Cartographer \citep{cartographer} to perform simultaneous
localization and mapping (SLAM) from data from the IMU and 3D LiDAR\@.
We record speech with the Blue Yeti microphone and convert it into text using
the Google Speech-to-Text \citep{google-speech-to-text} API\@.
We use the Stanford Parser \citep{stanford-parser} to parse the text.
We use Python's \texttt{pyttsx3} library \citep{python-speech-synthesizer} to
synthesize speech through the laptop speaker.
We use \textsc{YOLOv3} \citep{redmon2018yolov3} to detect people in images
taken from the Axis camera.
We use LSD (Line Segment Detector) \citep{GromponeVonGioi2010LSD:Control} on
images from the Axis camera as part of our method for detecting doors.
We extract text from images of door tags taken with the Axis camera with
Google's Optical Character Recognition \citep{googleOCR} API\@.
We use a LiDAR-camera calibration software package
\citep{lidarCameraCalibration} to compute a calibration matrix between the 3D
LiDAR and the Axis camera.
This matrix lets the system compute 3D locations for detected people and
doors in the environment.

\subsection{Architecture}

Our architecture is a finite-state machine illustrated in
\figref{fig:brain-fsm}.
This design makes logical steps towards reaching the goal and recovers from
failure at any of those steps.
Each state has a success condition that leads to the next state in the pipeline
as well as failure conditions which result in a transition to the initial state
of the system.
Given a goal description, the initial state is \wander, as the robot needs to
find a person to get directions to the goal.
This state allows the robot to explore the environment while simultaneously
attempting to detect and track people.
Once a person is found, the robot enters the \approachperson\ state, in which
it drives towards the person and synthesizes speech to grab their attention.
If the robot successfully reaches the person, it initiates a conversation with
them in the \holdconversation\ state.
The robot uses speech synthesis and speech recognition to request directions to
the desired goal and interpret the person's response, respectively.
It can ask an appropriate clarification question if the interpreted directions
are incomplete.
Once a complete set of directions are determined through dialogue with the
person, a transition to the \followdirections\ state is made.
\followdirections\ executes each direction in order by continuously mapping the
environment to determine when to continue to the next instruction, such as
turning \texttt{left} when a \textsf{left} turn becomes available.
Successful completion of the directions implies that the robot is in the
hallway that contains the goal.
The robot then enters the \navigatedoor\ state which involves detecting doors
and driving up to them to inspect their door tags.
This will let the robot ultimately confirm arrival at the desired goal.

\begin{figure*}
  \centering
  \resizebox{\linewidth}{!}{\begin{tabular}{@{}cccc@{}}
    \includegraphics[width=0.2\linewidth]{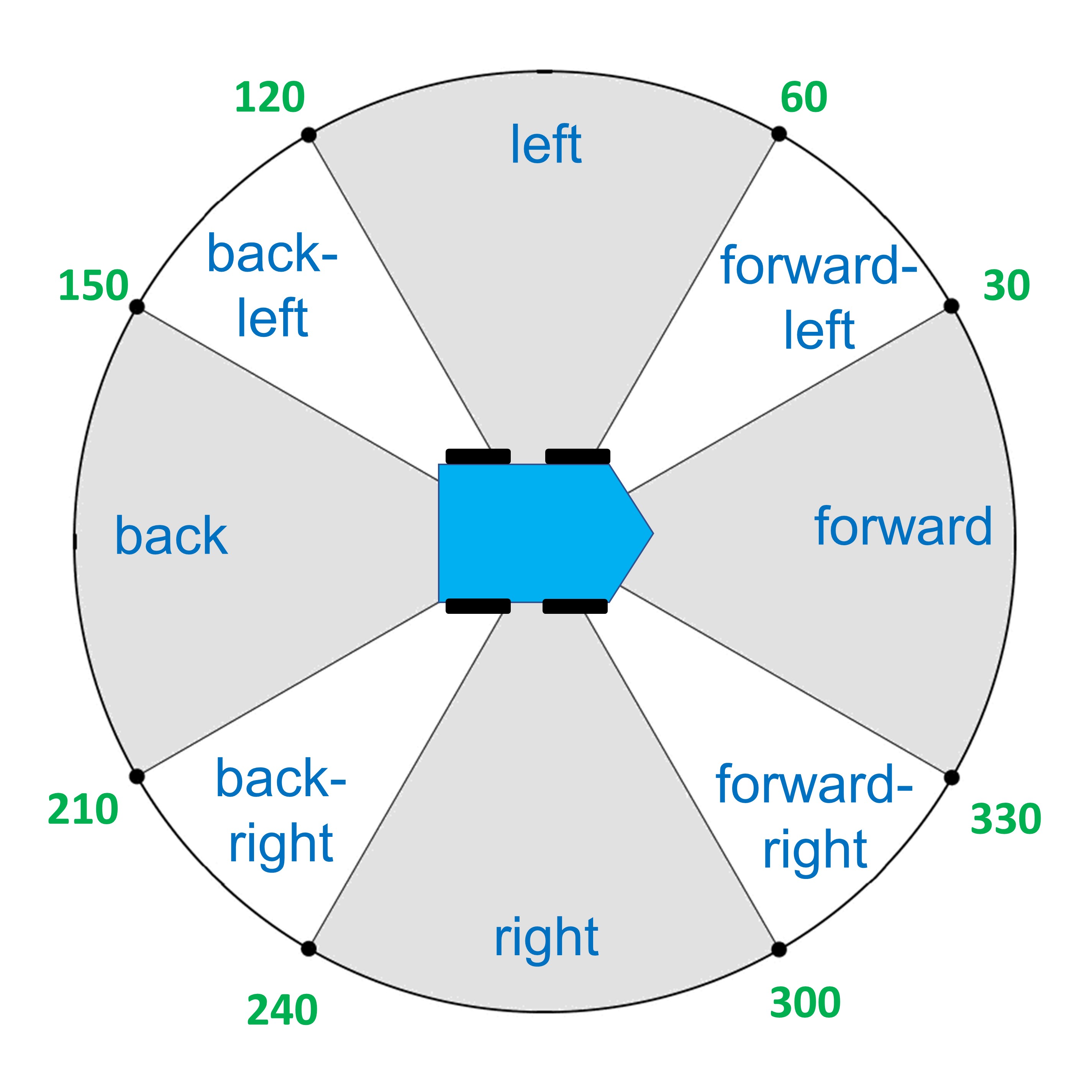}&
    \includegraphics[width=0.2\linewidth]{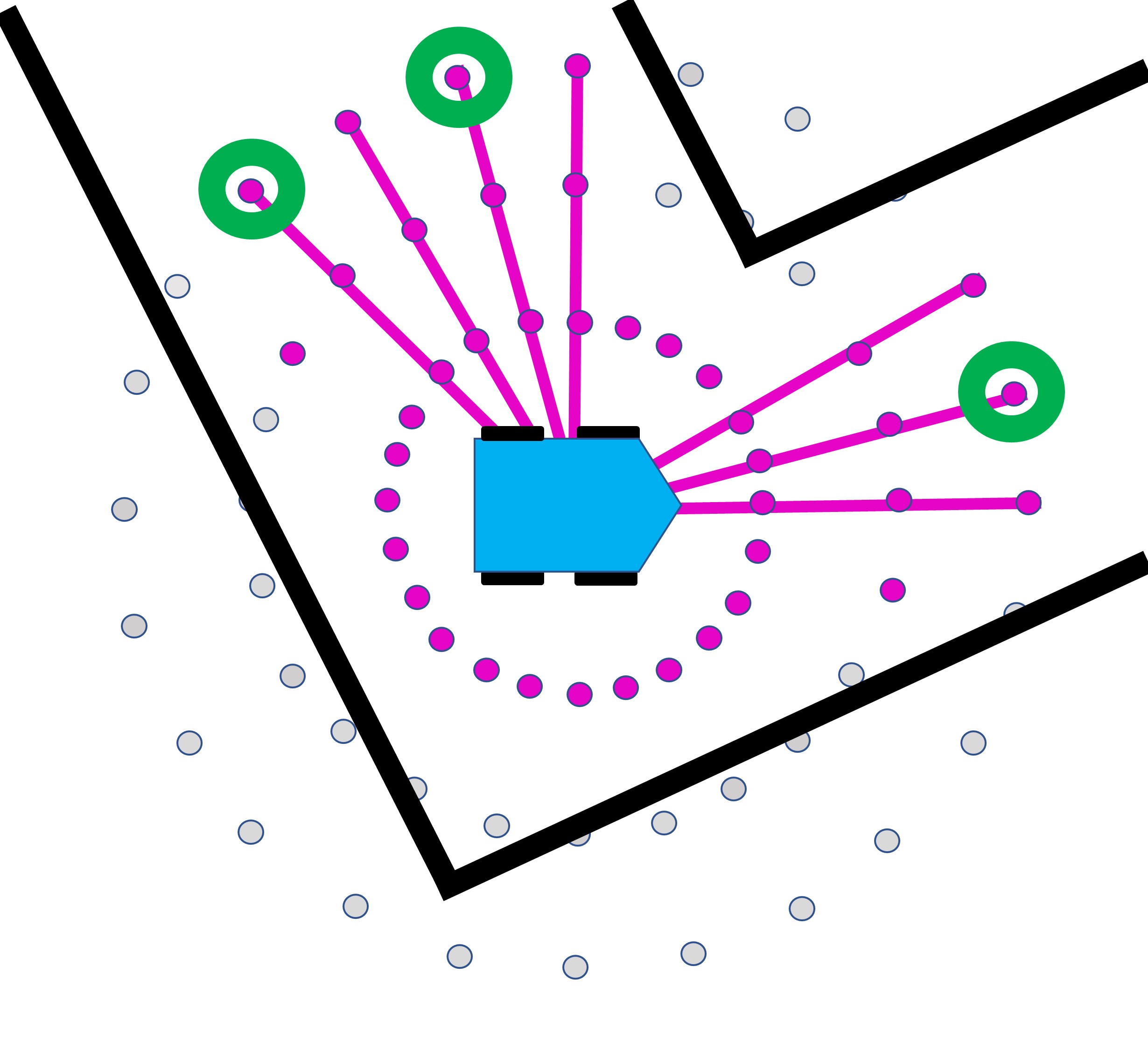}&
    \includegraphics[width=0.2\linewidth]{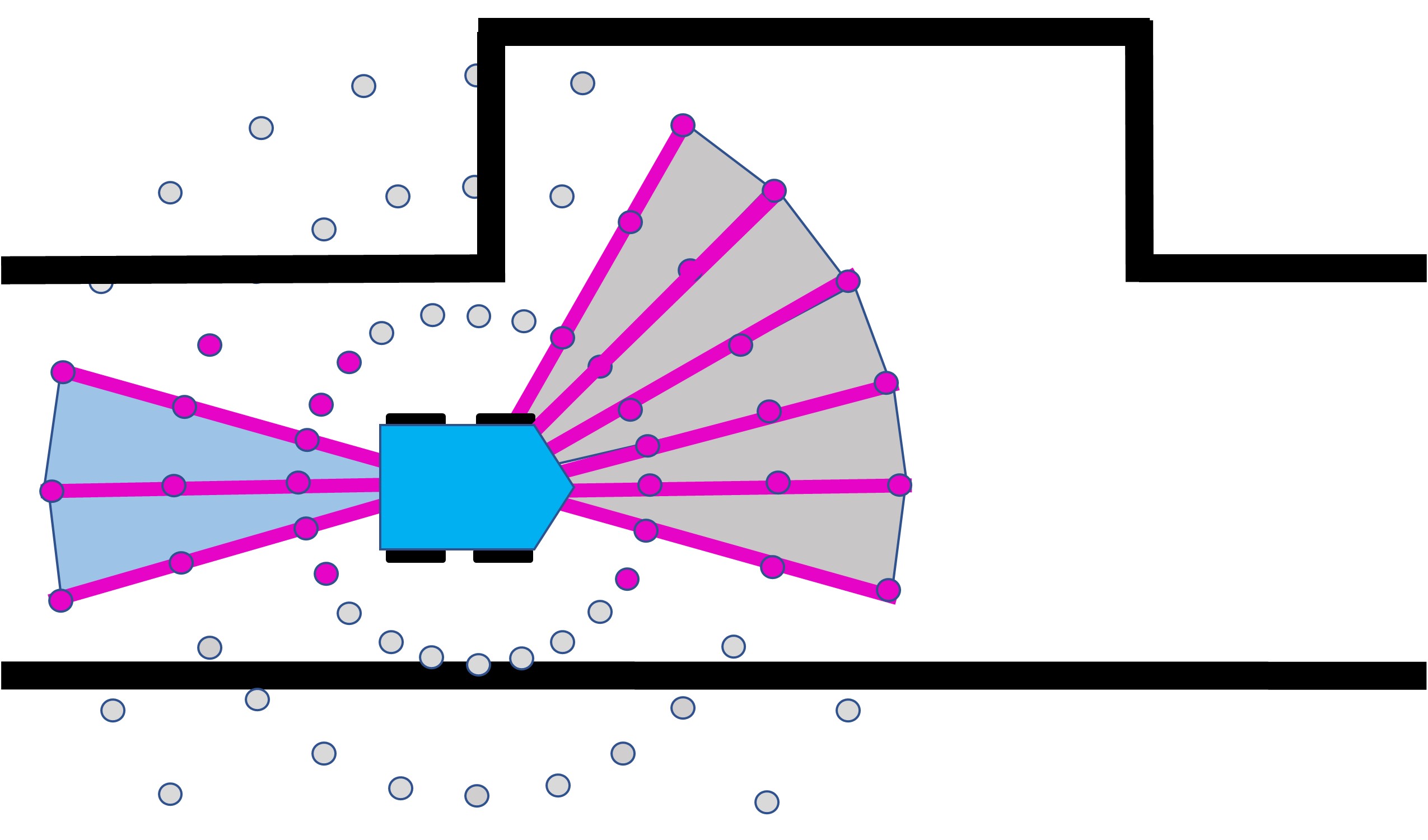}&
    \includegraphics[width=0.2\linewidth]{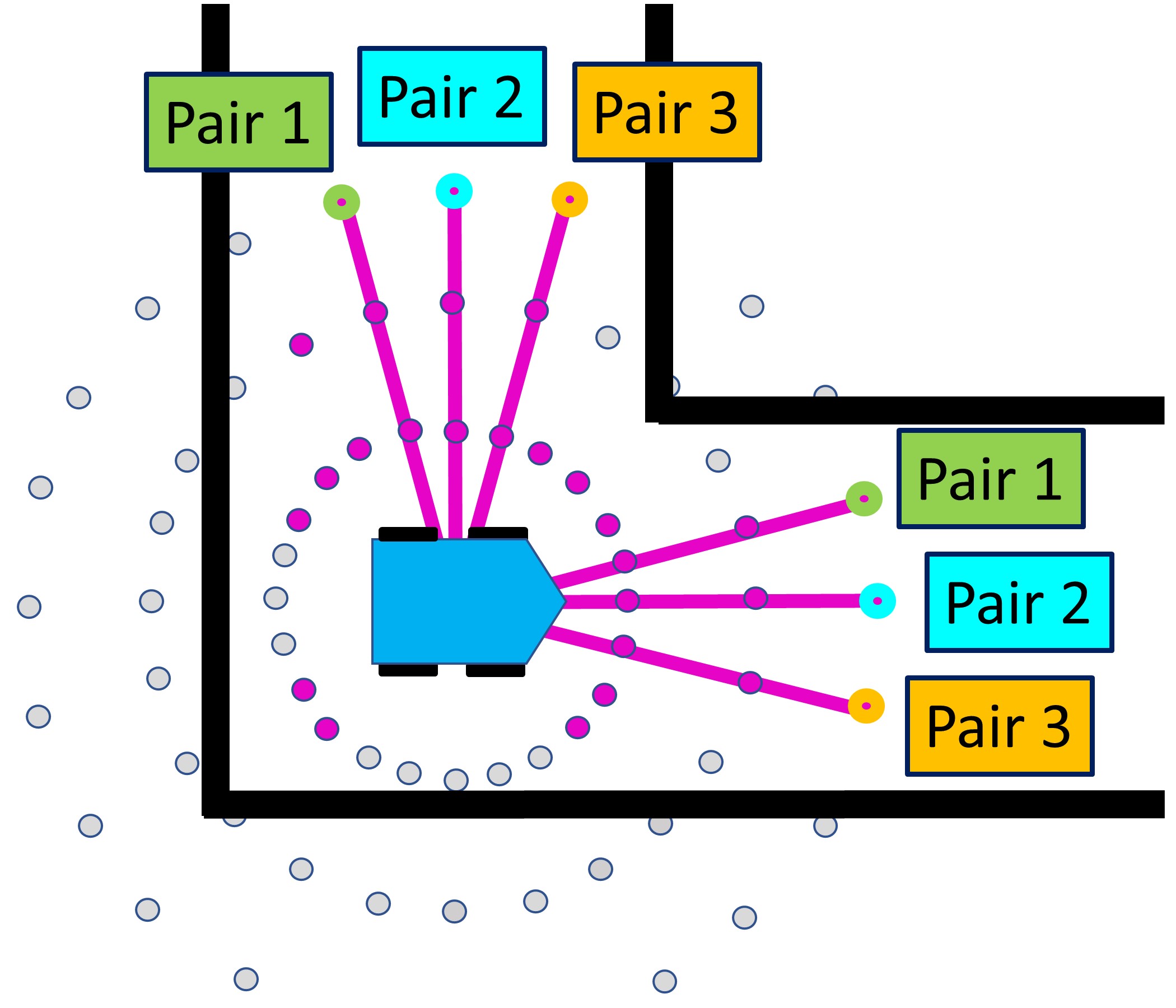}\\
    (a)&(b)&(c)&(d)
  \end{tabular}}
  \caption{(a)~Qualitative direction categories and their heading relative to
    robot orientation.
    (b)~An example of trajectory generation with trajectory distance
    limited to 3.6~m.
    The black lines represent obstacles (\ie\ walls).
    Target points of potential trajectories for 24 headings are shown as
    small points.
    The gray target points are filtered out as either unknown
    space or too near an obstacle.
    The pink target points represent drivable trajectories.
    The dark pink lines are headings associated with maximal drivable
    trajectories of distance 3.6~m.
    The dark green circles represent qualitative drivable trajectories with
    the qualitative direction labels \textsf{forward}, \textsf{left}, and
    \textsf{back-left}.
    (c)~An example of how intersection detection and classification
    ignores large traversable areas that arise from alcoves and other large
    open spaces.
    A single scale, namely a distance of 3.6~m, is depicted.
    The gray traversable area is much wider than a hallway so the drivable
    trajectories in that traversable area are discarded.
    Only the drivable trajectories in the blue traversable area are taken as
    hallway trajectories.
    (d)~An example of how intersection detection and classification, at
    a single scale of 3.6~m, forms candidate intersections as tuples of
    hallway trajectories at a given scale and selects at most one of these
    as the single detected and classified intersection.
    The target points of the hallway trajectories in Pair 2 are furthest
    from obstacles, so it will be selected.}
  \label{fig:navigation-process}
\end{figure*}

\subsection{Navigation Process}
\label{sec:navigation-process}

The physical environment has intersections connected by hallways.
Humans give navigation instructions by describing paths through hallways
between intersections.
They use informal terms to classify intersections into various types
(\eg\ \textbf{elbow}, \textbf{three-way}, and \textbf{four-way}) and to
distinguish between different hallways emanating from intersections by their
heading (\eg\ \textsf{forward}, \textsf{left}, and \textsf{right}).
To facilitate the interpretation of these informal navigation instructions,
our robot constructs and maintains two maps of the environment, one
quantitative and one qualitative.
The \emph{quantitative map} is an occupancy grid constructed by SLAM\@.
It is a 2D matrix of cells, where each cell corresponds to a 5~cm square of the
ground that takes one of three possible classifications: \emph{free},
\emph{obstacle}, or \emph{unknown}.
Generally, in an indoor office environment, free cells would be rooms or
hallways, obstacle cells would be walls or objects, and unknown
cells would be unexplored areas of the building or anything outside of the
walls.
The \emph{qualitative map} is a graph whose vertices represent detected
intersections labeled with intersection type and whose edges denote detected
hallway paths.
We refer to these as \emph{registered} intersections and hallway paths since the
process of constructing and maintaining the qualitative map can add, remove,
merge, and update registered intersections and hallway paths when detecting new
ones.
The quantitative map, along with the robot's current pose (position and
orientation) in that map, is continually maintained and updated by a background
process running SLAM\@.
The qualitative map is constructed and updated from the quantitative map and
robot pose by a background process continually running at 1~Hz.
We refer to the latter background process as the \emph{navigation process}.
The qualitative map produced by the navigation process is used by several of
our robot behaviors.

\subsubsection{Trajectory Generation}

The first step of the navigation process is to construct various sets of
trajectories, short paths that the robot can drive from its current pose.
A \emph{trajectory} is a target point in world coordinates at a specified
distance and heading from the current pose.
Low-level robot navigation uses trajectory target points as driving
instructions.
We nominally consider all distances that are integral multiples of 1.2~m from
1.2~m to 7.2~m, combined with all headings that are integral multiples of
$\frac{\textrm{360\degree}}{64}$, as \emph{potential trajectories}.
These are filtered as follows.
We first remove all trajectories which require traversing a point that is
within 0.6~m of an obstacle or unknown space to reach the target
point.
This yields a set of \emph{drivable trajectories}.
The 0.6~m threshold was chosen as it is our robot's circumscribed radius;
it would not fit through passageways smaller than this.
Since we need to search each trajectory for obstacles, the 1.2~m
quantization was chosen to reduce the number of potential trajectories
considered while still considering sufficiently many to successfully interpret
and execute human navigation instructions.
We then filter the set of drivable trajectories, keeping only the ones with the
largest distance for each heading.
This yields a set of \emph{maximal drivable trajectories}.
Finally, we label each maximal drivable trajectory with one of eight
\emph{qualitative directions} based on its heading
(\figref{fig:navigation-process}a).
We then filter the set of maximal drivable trajectories, keeping at most a
single trajectory for each qualitative direction, the one with the median
heading.
This yields a set of \emph{qualitative drivable trajectories}, there being at
most eight of these, each labeled with a distinct qualitative direction.
\Figref{fig:navigation-process}(b) illustrates the process of
trajectory generation.\footnote{While we classify eight distinct qualitative
  directions and qualitative driving directions, the remainder of the
  manuscript only considers the four primary ones.}

\subsubsection{Intersection Detection and Classification}

\begin{figure}
  \centering
  \resizebox{\linewidth}{!}{\begin{tabular}{@{}cc@{}}
    \includegraphics[width=0.5\linewidth]{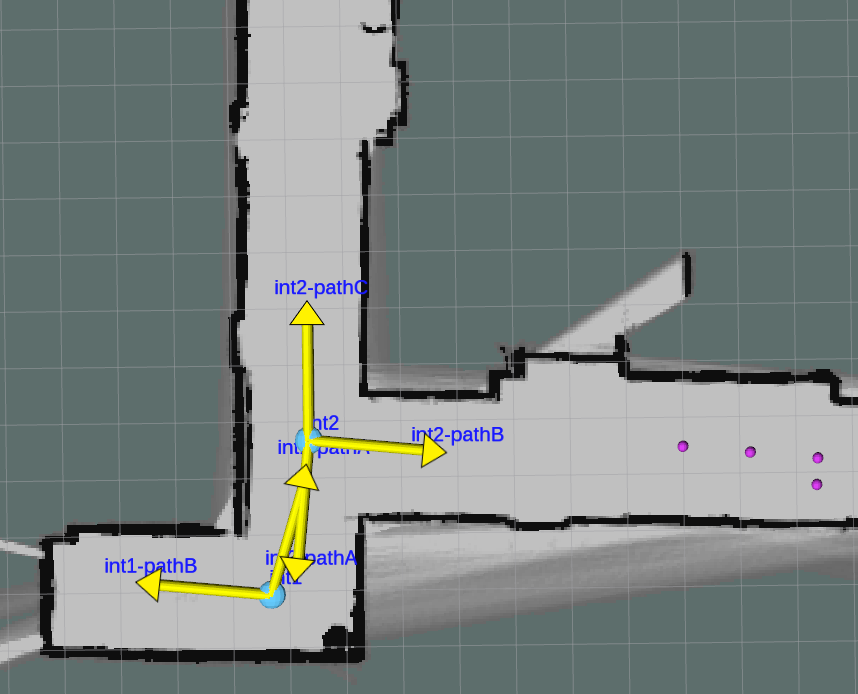}&
    \includegraphics[width=0.5\linewidth]{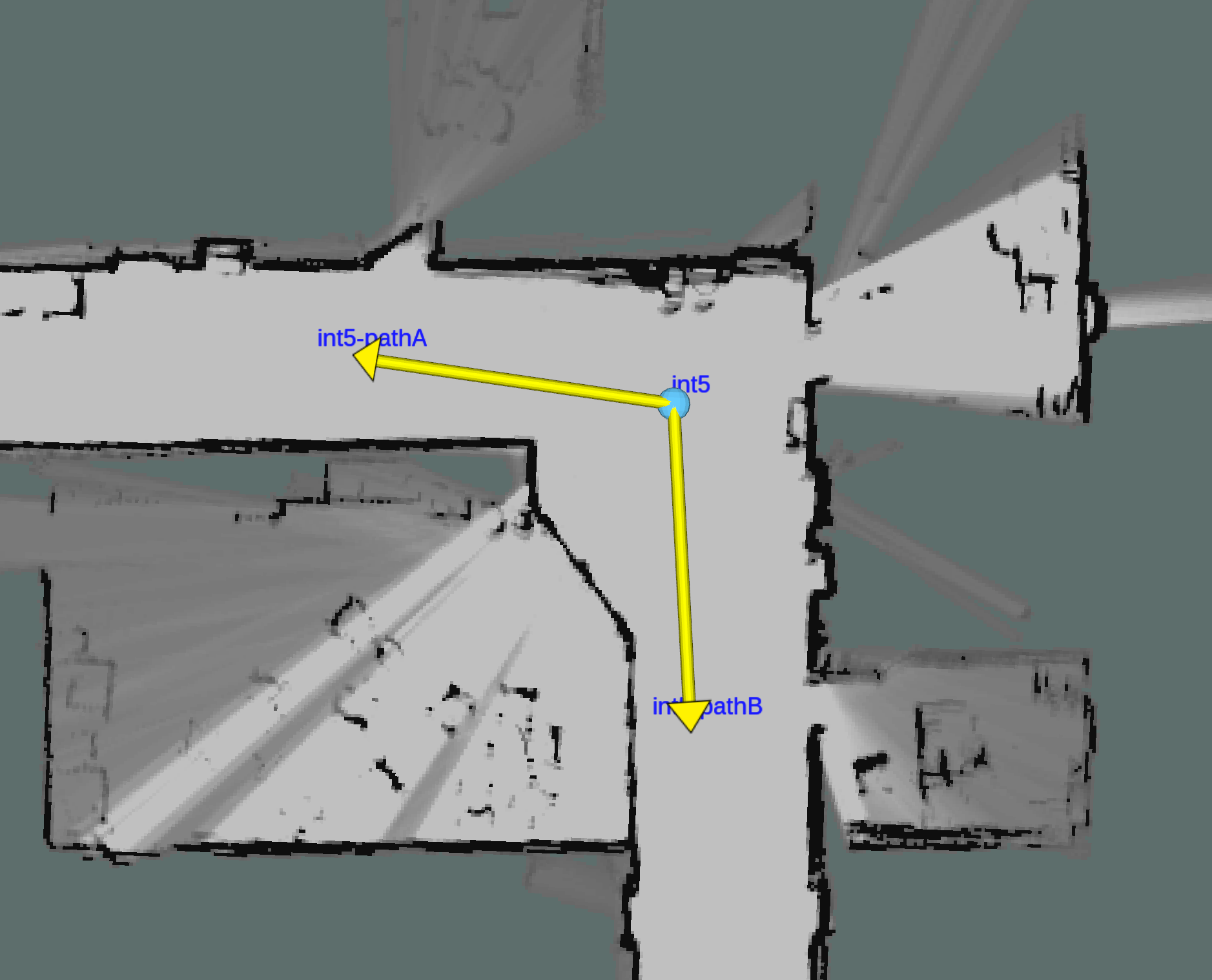}\\
    (a)&(b)
  \end{tabular}}
  \caption{(a)~These intersections, with very short hallways, can only be
    detected with a distance $\leq$2.4~m.
    (b)~This corner, with a decorative front to a lab space, can only be
    detected with a distance $\geq$4.8~m.
    All intersections are correctly detected and classified by considering
    multiple distances.}
  \label{fig:multi-scale-int-dets}
\end{figure}

The next step in the navigation process is to determine whether the robot is in
an intersection, and if so, to determine the type of that intersection
(\eg\ \textbf{elbow}, \textbf{three-way}, or \textbf{four-way}).
It does this at multiple scales, with each distinct trajectory distance taken
as a scale, to tolerate different building designs with different
hallway lengths and widths (\figref{fig:multi-scale-int-dets}).

At each scale, this process starts with all drivable trajectories at that
scale.
The first step is to eliminate drivable trajectories that would not be
considered as driving through hallways.
This is done by grouping all adjacent drivable trajectories (with headings
differing by $\frac{\textrm{360\degree}}{64}$) to represent a \emph{traversable
  area} (\figref{fig:navigation-process}c).
The width of a traversable area is determined as the maximal Euclidean distance
between any two target points of drivable trajectories in the traversable area.
If a traversable area is wider than the specified width of a hallway for that
building, all trajectories associated with that traversable area are discarded.
This allows the robot to ignore alcoves and similar open spaces, and yields a
set of \emph{hallway trajectories}.

The hallway trajectories are grouped into pairs, triples, and quadruples
whose headings are $\approx$90\degree\ apart.
These constitute potential intersections of type \textbf{elbow},
\textbf{three-way}, and \textbf{four-way}, respectively.
Each such tuple constitutes a \emph{potential intersection}.
Potential intersections whose elements are not separated by obstacles are
discarded.
This yields a set of \emph{intersection candidates}, each of which has the
robot's position as its initial location.
Intersection candidates are scored based on the sum of the distances of the
location and each element's target point from the nearest obstacle.

%\needswork: the distance of the robot position from obstacles is scored for
% refinement

This process may produce multiple intersection candidates, \eg\ an
\textbf{elbow} will always be present when a \textbf{three-way} is also present
and an intersection candidate produced at a larger scale may also be present at
a smaller scale.
The navigation process prioritizes intersection candidates as follows:
\begin{inparaenum}
\item quadruples over triples,
\item triples over pairs,
\item larger scales over smaller scales, and
\item higher scores over lower scores.
\end{inparaenum}
This yields at most a single detected and classified intersection at the current
robot position.
For reasons to be discussed later, each detected and classified intersection is
given a unique identifier and each hallway trajectory in that intersection tuple
is also given a unique identifier.

As an example, consider \figref{fig:navigation-process}(d), which illustrates
intersection detection and classification at a single scale, namely 3.6~m.
Three pairs of hallway trajectories (green, blue, and orange) are shown (out of
a possible nine), each hallway trajectory associated with the highlighted
(green, blue, and orange) target point.
Each such pair constitutes an intersection candidate.
Pair 2 (blue) will have the highest score (since its target points are furthest
from obstacles) and will be kept as the single detected and classified
intersection.

\subsubsection{Intersection Refinement}

Since the navigation process performs intersection detection and classification
repeatedly at 1~Hz, it can detect the same intersection multiple times,
particularly if the robot remains in an intersection for more than 1~s.
This is akin to how object detectors can place multiple but different boxes
around an object.
We address this by performing a kind of nonmaximal suppression (NMS) in a
fashion analogous to how object detectors deal with this problem.
Moreover, intersection detection and classification is performed relative to
the robot pose.
As this changes, the classified intersection type might change due to noise
(\eg\ classifying an intersection as an \textbf{elbow} when it is a
\textbf{three-way}).
Further, through a process described below, detected and classified
intersections are registered as vertices in the qualitative map.
Registered intersections are labeled with their location, which we wish to be
the center of the physical intersection.
However, detected intersections are registered with their location being the
robot position at the time of detection, which might not be the center of the
physical intersection.

To address this, we perform intersection refinement.
We suppress detection of new intersections when the robot is within 2~m of a
registered intersection.
Further, we continuously refine the locations and classification labels of
registered intersections in a background process.
This process recomputes the intersection candidates while imagining the robot
to be at every point in a 3$\times$3 grid, with 0.4~m spacing, centered on the
current location of each registered intersection, using the current SLAM
occupancy grid.
The intersection candidates are pooled and prioritized as in initial
intersection detection and classification to yield a single redetection and
reclassification.
The registered intersection is updated to reflect the redetected and
reclassified intersection, including its location.
We enforce a constraint that intersections cannot shift more than 2.4~m from
their original location.
For efficiency, this refinement is only applied to a registered intersection
when the robot is within 5~m of that intersection.

Each detected and registered intersection contains
\begin{inparaenum}
\item its location,
\item its type, and
\item a tuple of hallway trajectories.
\end{inparaenum}
The hallway trajectories represent the hallway paths between intersections.
During refinement, the redetected intersection might contain different hallway
trajectories than the current registered intersection.
This can happen when, for example, a group of people stop to talk and block a
hallway trajectory in the intersection.
During intersection refinement, we wish to maintain the same unique identifiers
associated with each intersection and hallway trajectory in that intersection.
This requires constructing a correspondence between the hallway trajectories in
the registered intersection and those in the redetected intersection.
The optimal correspondence is found with the Hungarian algorithm
\citep{kuhn1955hungarian} applied to a bipartite graph whose left vertices are
the hallway trajectories in the registered intersection, whose right vertices
are the hallway trajectories in the redetected intersection, and whose edges are
given a cost which is the angular distance between the hallway trajectory
headings.
Edges in this bipartite graph are only created when the angular distance is
$<$30\degree.
This correspondence is used to reassign the unique identifiers from the
hallway trajectories in the registered intersection to those in the redetected
intersection.

As described previously, a redetected intersection may have different type than
a registered intersection, and thus may have a different number of hallway
trajectories.
Thus, the correspondence as produced above may fail to assign a hallway
trajectory from the registered intersection to one in the redetected
intersection either because the redetected intersection is of a different type
with fewer hallway trajectories or because of eliminated edges.
A hallway trajectory in the registered intersection that does not have a
corresponding hallway trajectory in the redetected intersection is maintained,
with its unique identifier, in the registered intersection in a
\emph{deactivated} state, so that it can be reactivated later with the same
unique identifier.

\subsubsection{Intersection Graph}

\begin{figure}
  \centering
  \includegraphics[width=\linewidth]{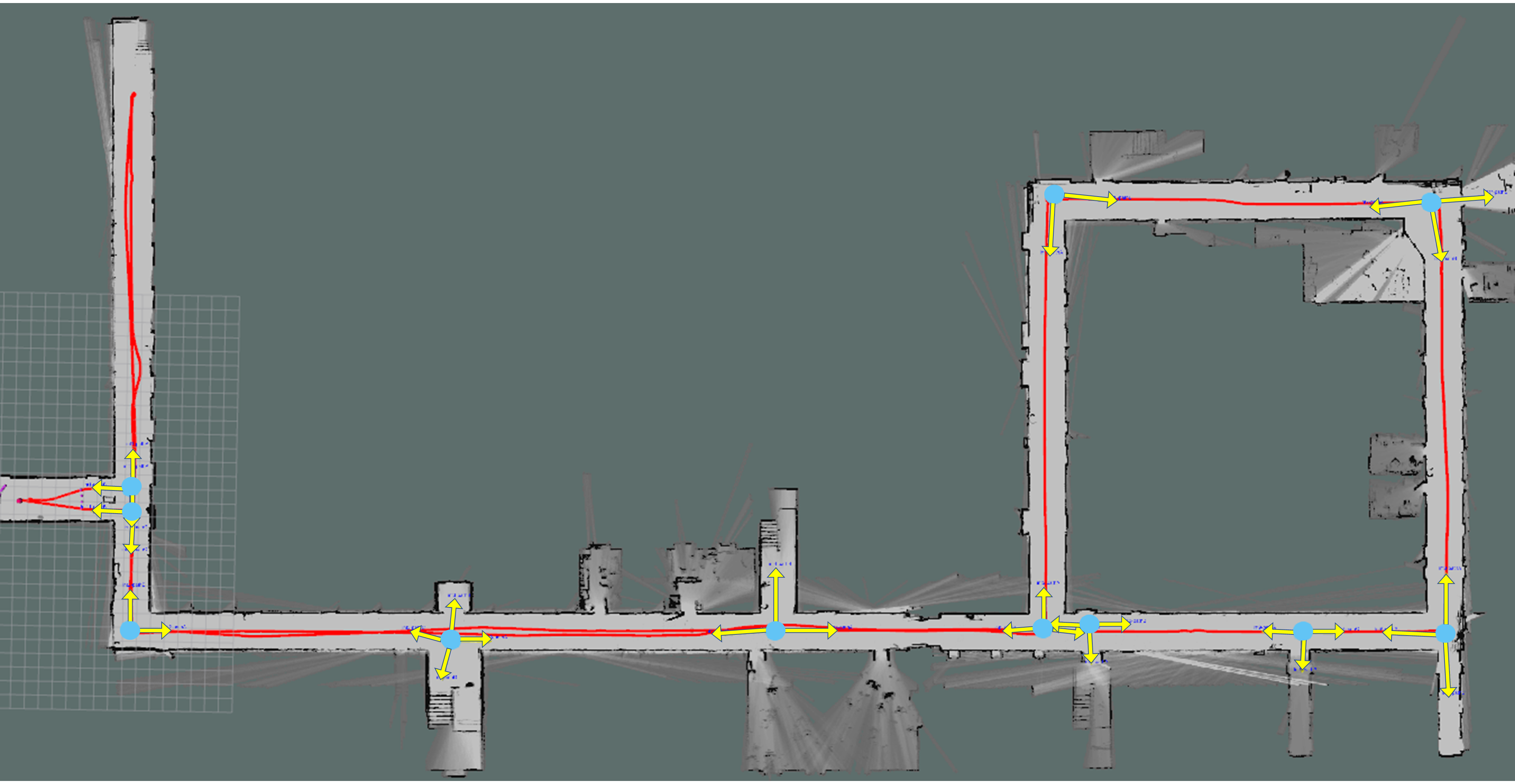}
  \caption{Qualitative map produced for a single floor of a building.
    The red line indicates the path that the robot traveled.
    Blue spheres are registered intersections.
    Yellow arrows are hallway trajectories associated with each registered
    intersection.}
  \label{fig:phys}
\end{figure}

The unique identifiers associated with registered intersections and their
hallway trajectories, whose consistency is maintained over time by intersection
refinement, allow us to construct a graph to represent the qualitative map.
When the robot comes within 0.5~m of a registered intersection, it selects the
hallway trajectory whose target point is closest to the robot position as the
one used to enter the intersection.
The robot's path is searched backward until it comes within 3~m of a hallway
trajectory from the last registered intersection it visited.
An edge in the qualitative map is registered between these two registered
intersections and represents a hallway path.
The weight of this edge is taken as the Euclidean distance between
the locations of the two registered intersections.
If the backward search process does not produce a hallway trajectory, that
hallway trajectory is left unconnected, to represent hallways that have not been
completely explored yet.
This graph is continually constructed and updated by the navigation process.
See \figref{fig:phys} for an example of the qualitative map produced for a
single floor of a building.

\section{States of the System Architecture}

In this section, we describe the the implementation of each state in our system.

\subsection{\wander}

\wander\ is the initial state of the system and lets the robot
continuously navigate through the environment, trying to detect and track
people until it finds an approachable person.
The general strategy is to explore the environment in a human-like manner,
usually moving forward, choosing a direction to go when at an intersection, and
only turning around when reaching a dead end.

\subsubsection{\wander\ Substates}
\label{sec:wander-fsm}

The \wander\ state has five substates:
\begin{inparaenum}
\item \makedecision,
\item \rotaterecovery,
\item \rotate,
\item \driveforward, and
\item \drivethroughintersection.
\end{inparaenum}
\wander\ enters the \makedecision\ substate first, where it analyzes whether
it is in a registered intersection and which qualitative directions are
available.
If no qualitative directions are available (\eg\ when it is first initialized),
it enters the \rotaterecovery\ substate, which causes it to spin in place
360\degree.
This substate helps to update the quantitative and qualitative maps in the
immediate vicinity, which determine whether any qualitative drivable
trajectories are available.
If none are available, it stays in this substate.
If qualitative drivable trajectories are available, the robot selects the one
whose heading is closest to its current orientation as the \emph{recovery
  angle}.
The robot enters the \rotate\ substate to match its orientation with that of
the recovery angle and will then enter the \driveforward\ state.
This series of substates helps it find its way out of alcoves, entrances, or
elevator landings and into hallways.

When in the \driveforward\ state, the robot continuously drives
\emph{\textbf{forward}} while monitoring for registered intersections.
This \emph{\textbf{forward}}, which we use to indicate when and how a robot
drives down a hallway, is distinct from the qualitative drivable trajectory
\textsf{forward} and will be explained in \secref{sec:forward-goals}.
When it enters a registered intersection, the robot enters the
\makedecision\ substate to determine what to do.
First, it determines which hallway trajectory it entered the registered
intersection from.
Hallway trajectories included in registered intersections maintain a
\emph{visitation time} as an indication of when they were last visited.
The visitation time for the hallway trajectory used to enter the registered
intersection is updated to the current time.
Then, it temporally orders all active hallway trajectories of that
registered intersection.
If it has visited each hallway trajectory, it selects the oldest one and takes
it.
If there are one or more hallway trajectories that it has not taken, it randomly
selects one and takes it.
It then updates the visitation time of the selected hallway trajectory with the
current time.
Maintaining visitation times allows the robot to explore the environment in a
thorough manner, preferring to visit areas in the map that it has not seen or
that it has seen least recently.

\begin{figure*}
  \centering
  \begin{tabular}{@{}c@{}c@{}}
    \begin{subfigure}[b]{0.5\linewidth}
      \includegraphics[width=\textwidth]{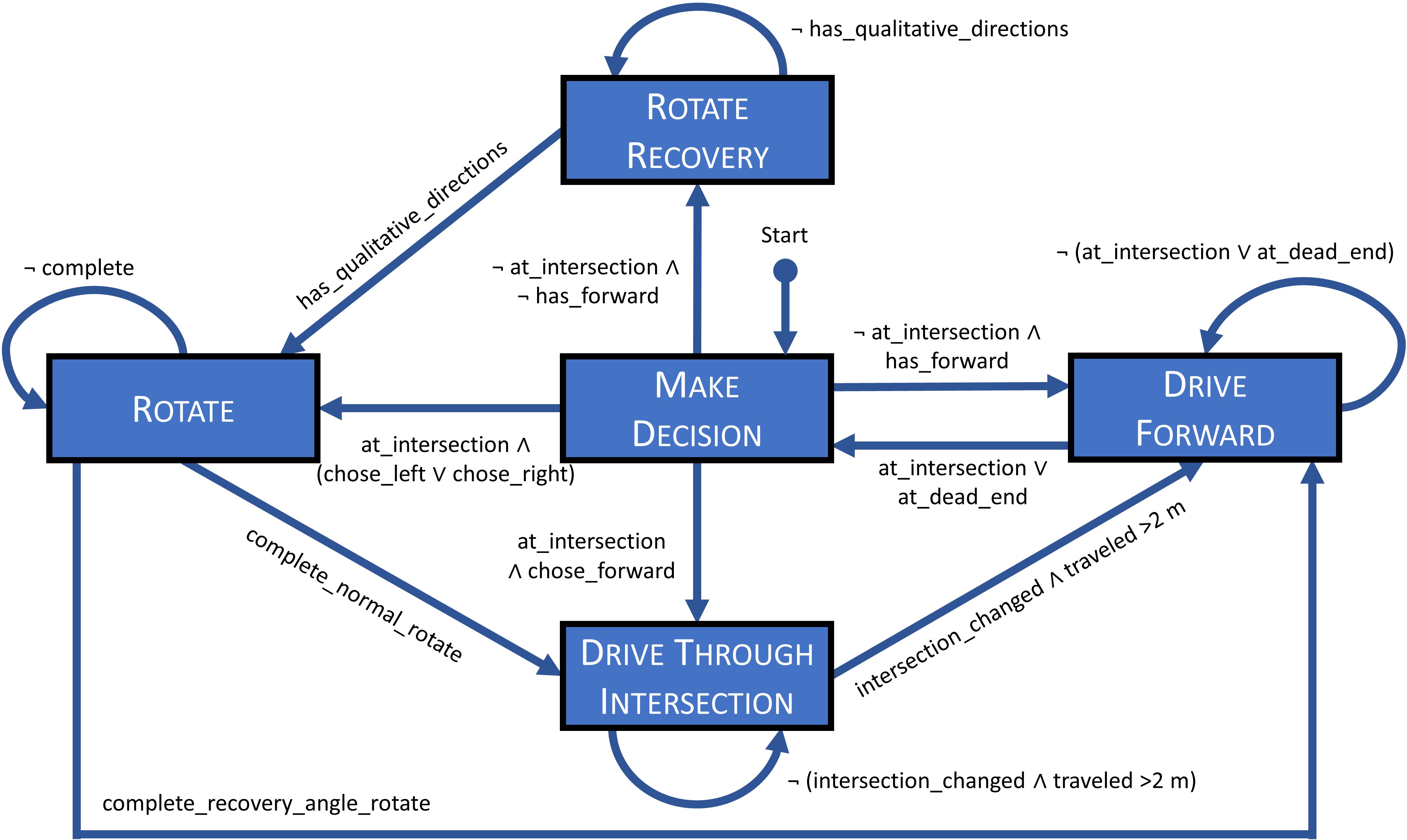}
      \caption{The \wander\ finite-state machine.}
      \label{fig:wander-fsm}
    \end{subfigure}&
    \begin{subfigure}[b]{0.5\linewidth}
      \includegraphics[width=\textwidth]{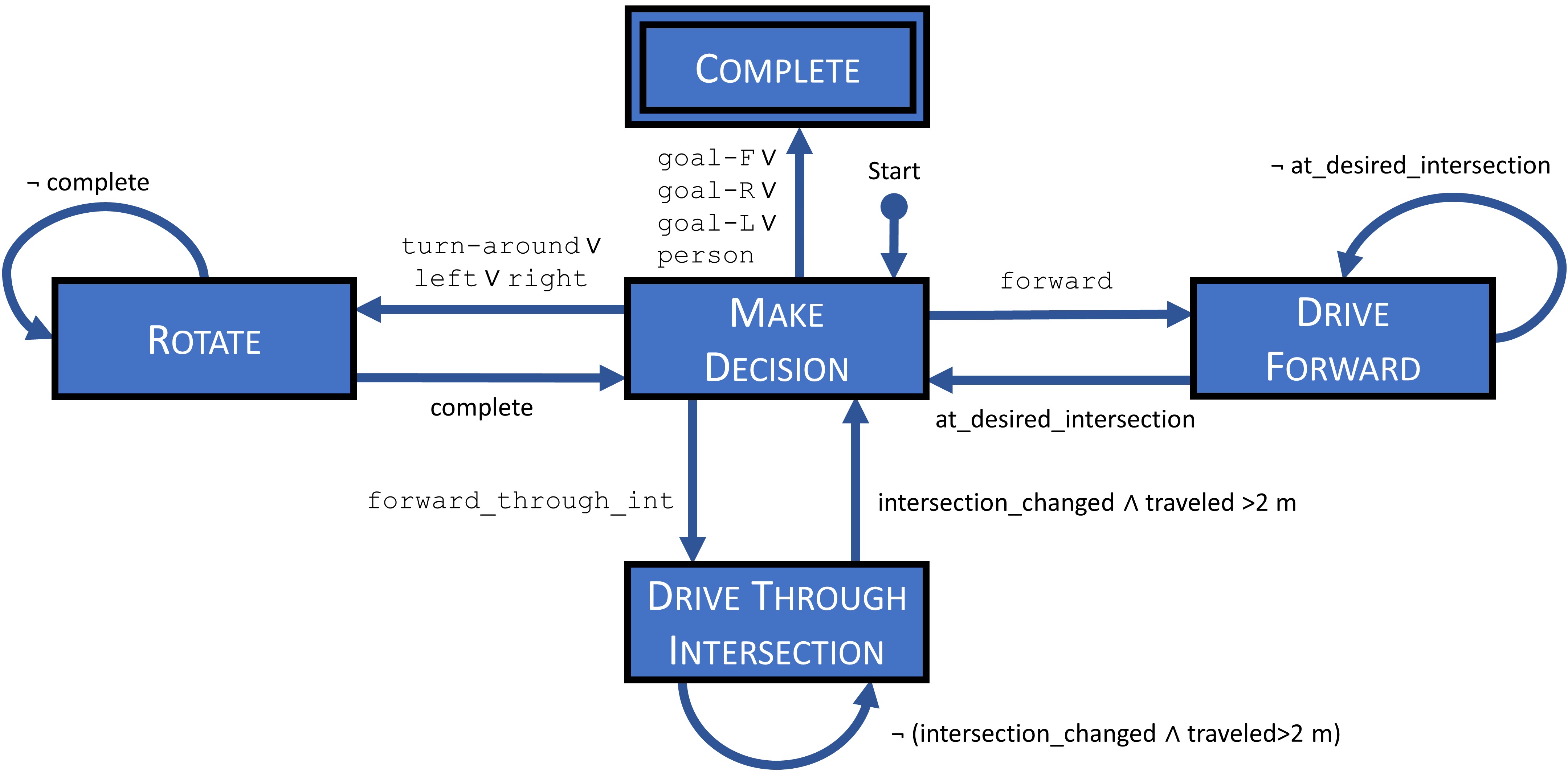}
      \caption{The \followdirections\ finite-state machine.
        Transition conditions in \texttt{Courier} represent the next step in the
        plan.
        All others represent transition conditions derived from sensor data.}
    \label{fig:follow-directions-fsm}
    \end{subfigure}
  \end{tabular}
  \caption{Finite-state machines internal to the \wander\ and
    \followdirections\ states.}
  \label{fig:subfsms}
\end{figure*}

After selecting the hallway trajectory to take, if the hallway trajectory is to
the \textsf{left} or \textsf{right} of the robot's pose, the robot will enter
the \rotate\ substate to rotate 90\degree\ to the left or right, respectively.
Thereafter, or in the case that the robot chose to drive straight through the
intersection, it enters the \drivethroughintersection\ substate to move out
of the intersection and into the hallway.
This substate prevents the robot both from immediately recognizing it is in a
registered intersection and causing it to re-evaluate what to do.
The \drivethroughintersection\ substate has the robot drive
\emph{\textbf{forward}} continuously.
Once it has traveled more than 2~m and the intersection type has changed, the
robot returns to the \driveforward\ substate.

If, while driving \emph{\textbf{forward}}, the robot reaches a dead-end (which
is determined by there being a \textsf{back} qualitative drivable trajectory
but no \textsf{forward}, \textsf{left}, or \textsf{right} qualitative drivable
trajectories), the robot performs a single 360\degree\ spin to ensure that it
truly is at a dead-end (instead of simply having failed to detect an adjoining
hallway).
After performing this spin, it will enter the \makedecision\ substate, which, if
it is still at a dead-end, will enter the \rotaterecovery\ substate.
This substate will see that there is a single qualitative drivable trajectory
available (\textsf{back}), which it uses as its recovery angle.
The robot would enter the \rotate\ substate, rotate 180\degree, and then return
to the \driveforward\ substate.
These substates can be modeled as a FSM as shown in \figref{fig:wander-fsm}.

\subsubsection{Forward Driving Goals}
\label{sec:forward-goals}

When the robot starts driving down a hallway, it might not be in the middle of
the hallway, and its orientation might not match that of the hallway.
Despite starting in a suboptimal pose, we want the robot to move towards and
drive down the middle of the hallway.
To do so, when the robot enters the \driveforward\ substate, we take its
current orientation as the \emph{\textbf{forward} orientation}.
This \emph{\textbf{forward}} orientation will get refined over time as the robot
drives down the hallway to more accurately reflect the same orientation as the
hallway.

While the navigation process provides a set of qualitative drivable
trajectories, due to the suboptimal starting pose of the robot, there may not
be a qualitative drivable trajectory associated with the \emph{\textbf{forward}}
orientation.
Therefore, we pass the \emph{\textbf{forward}} orientation and a \emph{cone
  angle} to the navigation process, ask it to find all maximal drivable
trajectories within that cone and return the median one as the \emph{median
  drivable trajectory}.
Initially, the cone angle is $\pm\textrm{15\degree}$ from the
\emph{\textbf{forward}} orientation.
If no drivable trajectories are found within the cone, this process is repeated
with a cone angle of $\pm\textrm{30\degree}$ and then $\pm\textrm{45\degree}$.
When drivable trajectories are found within the cone, the median drivable
trajectory is used as the \emph{\textbf{forward} driving goal}.

\begin{figure}
  \centering
  \includegraphics[width=0.5\linewidth]{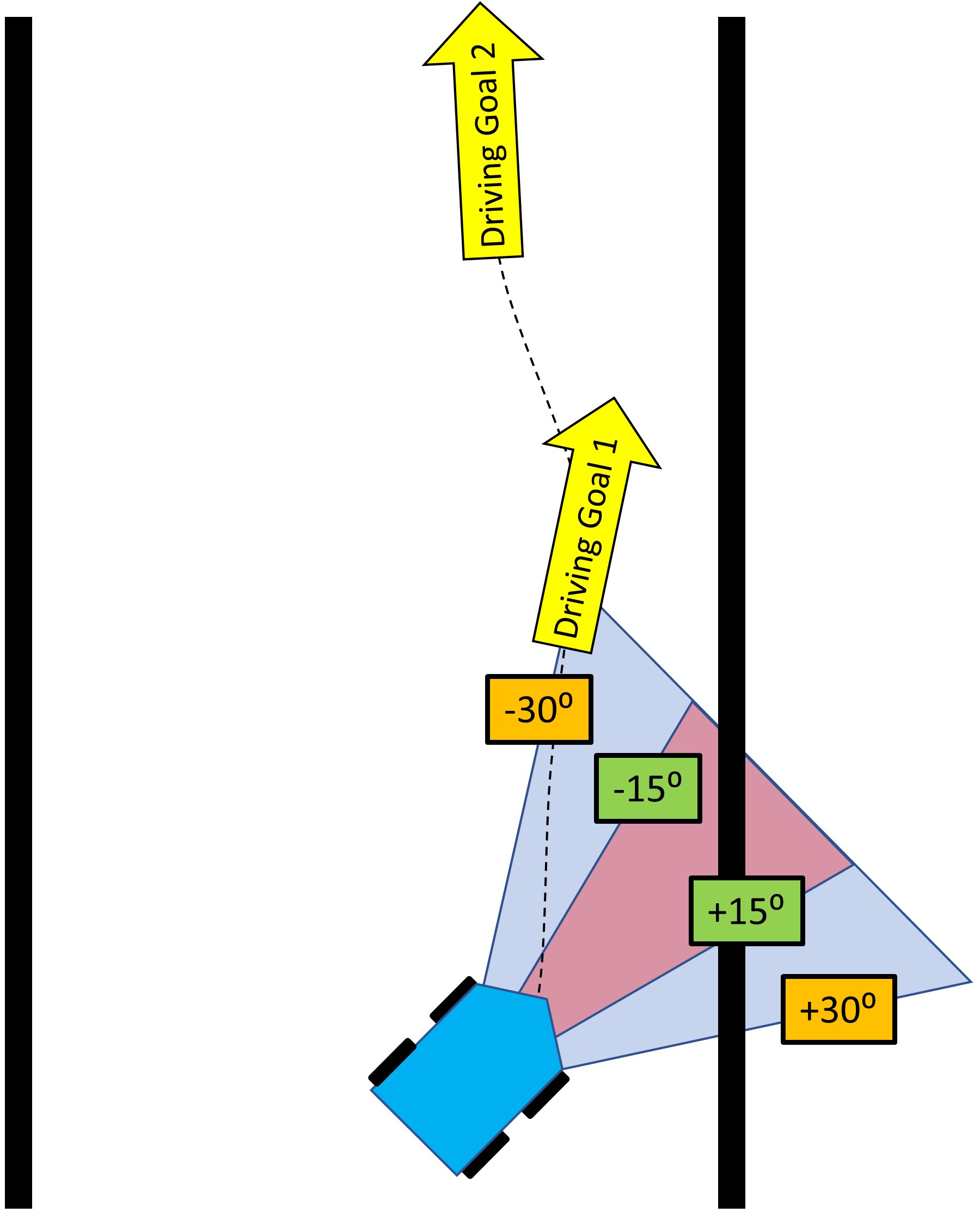}
  \caption{Example of how the robot determines a \emph{\textbf{forward}}
    driving goal despite starting with a suboptimal pose.
    The robot is 1~m from the wall and its orientation is
    $\approx$45\degree\ off from that of the hallway.
    The transparent red cone indicates that the navigation process was unable
    to find any drivable trajectories using a cone angle of
    $\pm\textrm{15\degree}$ from the \emph{\textbf{forward}} orientation.
    The transparent blue cone indicates that the navigation process was able to
    find a drivable trajectory using a cone angle of $\pm\textrm{30\degree}$ and
    thus a cone angle of $\pm\textrm{45\degree}$ is not required.
    The median drivable trajectory that the navigation process returned is
    used as a \emph{\textbf{forward}} driving goal (labeled ``Driving Goal 1'').
    Upon approaching that driving goal, the robot repeats this process, but
    only up to a cone angle of $\pm\textrm{30\degree}$ if necessary, and the
    yellow arrow labeled ``Driving Goal 2'' will be its second
    \emph{\textbf{forward}} driving goal.
    This process incrementally moves the robot to the center of the hallway and
    changes its orientation to more closely match that of the hallway.}
  \label{fig:forward-goals-wall}
\end{figure}

This gets the robot moving towards the middle of the hallway and having its
orientation more closely reflect that of the hallway.
As it nears this first driving goal, these steps are repeated, but only up to a
window of $\pm\textrm{30\degree}$.
After nearing its second driving goal, the robot is typically in the middle of
the hallway and has the same approximate orientation as the hallway.
For all remaining \emph{\textbf{forward}} driving goals after this point, only
a window of $\pm\textrm{15\degree}$ is used.
Decreasing the maximum possible window after navigation to the first and second
driving goals is done to avoid \emph{\textbf{forward}} driving goals that lead
to undesirable behavior.
As the robot drives down a hallway, there may be alcoves on the left or right.
If a wide cone angle is used, drivable trajectories can be within the alcove and
the median drivable trajectory that is returned from the navigation process can
cause the robot to veer off from the center of the hallway.
In the worst case, it can drive into the alcove, have no qualitative direction
\textsf{forward}, and think it has arrived at a dead-end.
Alternatively, if instead of there being an alcove, there is a hallway, the
robot can inadvertently drive around a corner (without realizing it) and still
think it is driving \emph{\textbf{forward}}.
See \figref{fig:forward-goals-wall} for an illustration of this entire process.

Earlier, we stated that the \emph{\textbf{forward}} orientation is updated over
time to more accurately reflect the same orientation as the hallway.
At first blush, it may seem logical to always use the robot's current
orientation as the \emph{\textbf{forward}} orientation.
However, this fails under the following scenario.
As the robot drives down a hallway, obstacles may appear in front of it
(\eg\ people walking and/or sometimes stopping in front of it---whether
unintentionally or intentionally---to see how the robot will react).
Standard local path planners handle these types of unexpected obstacles
by navigating around them.
As the robot is driving around such an obstacle, its orientation may
change significantly such that it no longer matches that of the hallway and in
some cases can be nearly perpendicular to the hallway.
If, at this moment, the robot needed to update its driving goal, and used its
current orientation as the \emph{\textbf{forward}} orientation, the navigation
process could either determine that there were no drivable trajectories or
return a drivable trajectory that would move the robot in a direction it should
not go.
Either could derail the robot from properly driving down the hallway.

To handle this case, once the robot has moved 2~m from its starting point, we
compute the angle from its starting point when it began driving down to the
hallway to its current position and use this as the \emph{\textbf{forward}}
orientation.
This gives a much more accurate approximation of the orientation of the hallway
and makes the \emph{\textbf{forward}} orientation robust to the situation
described above.

There is one additional mechanism we employ to help the robot drive
\emph{\textbf{forward}} around obstacles but we will postpone that discussion
until \secref{sec:forward-goals2} as it makes more sense in that context.

\ignore{
\subsubsection{Smooth Driving}

If a robot is continuously fed new driving goals, it has to pause momentarily to
re-plan.
This pause is extremely brief, but it causes the robot to ``jerk,'' resulting in
navigation that might alarm passersby.
On the other end of the spectrum, if the robot is allowed to complete driving
to its driving goal before it is issued a new driving goal, it will stop/start
many times as it drives down a hallway.
Furthermore, if the driving-goal pose isn't aligned with that of the robot's
path, it may rotate in place after reaching the driving goal.
These behaviors are unfamiliar and non-intuitive to non-trained humans.
To reduce the frequency of this ``jerking'' behavior and to facilitate smooth
driving, we do the following:
\begin{inparaenum}
\item we compute driving goals that are far ahead of the robot's current
  position (7.2~m) and
\item when we come within a 2.5~m of the driving goal, we issue a new driving
  goal.
\end{inparaenum}
These two things result in smoother driving, completely eliminating the
start/stop/rotate behavior and only requiring the momentary ``jerk'' every
$\approx$5.0m.}

\subsubsection{Person Detection and Tracking}

The ultimate goal of the \wander\ state is to locate a person to
approach.
Therefore, while the robot is navigating through the environment, it uses the
Axis camera and 3D LiDAR to continuously detect and track people in that
environment.

\textsc{YOLOv3}, a real time object-detection system, is used to detect people.
It can detect multiple objects of different classes and provide their spatial
location via bounding boxes.
These boxes are used to determine the location of people in the quantitative
map.

The camera's horizontal field of view angle and resolution are known.
Each pixel in the camera can be mapped to a particular angle relative to the
position of the camera.
These angles can be used to extract the distance of the detected object from the
LiDAR data.
Because bounding boxes from object detectors are imperfect, and because
standard object detectors localize detections with boxes even though the shape
of most natural objects (\eg\ humans) is not rectangular, some of the LiDAR
points will shoot beyond the object and hit a more distant object or wall.
Instead of taking the average of these points, we keep the closest one, which
represents the nearest part of the detected person.
By knowing the pose of the robot, camera, and LiDAR, we can determine the
precise position of that object in the quantitative map.

These person detections, along with their quantitative map coordinates, are fed
into a detection-based tracker.
Typically, a detection-based tracker combines bounding boxes from
adjacent image frames into tracks if their pixel coordinates sufficiently
overlap.
However, in our case, this method is unsuitable because the robot is constantly
moving.
Therefore, instead of combining detections that occupy the same spatial region
in the camera's field of view, we combine detections that occupy/overlap the
same location in the quantitative map.
In this way, if the robot moves or rotates, the moving field of view does not
affect the ability of the tracker to accurately piece together detections into
tracks.

For all existing tracks, we compute the Euclidean distance from the last known
location to each new detection.
These distances constitute the cost to pair the track with the detection.
We use the Hungarian algorithm to minimize the cost across all pairings of
tracks with new detections and find the optimal pairing.
We perform a sanity check to make sure that no pairings are unreasonable.
We first determine the walking speed of the person, by computing how far they
have traveled since the last detection and dividing that by the amount of time
that has elapsed since then.
If the walking speed is $>$2.5~m/s, we break the track-detection
pairing and the detection becomes a new track.
All pairings that meet this threshold are combined into a single track.
Because detectors can be noisy and imperfect, the tracker employs forward
projection on all tracks for up to 0.5~s to compensate for missed
detections.
Any tracks that go 0.5~s without an additional detection are pruned.

\begin{figure}
  \centering
  \resizebox{\linewidth}{!}{\begin{tabular}{@{}cc@{}}
    \includegraphics[height=100pt]{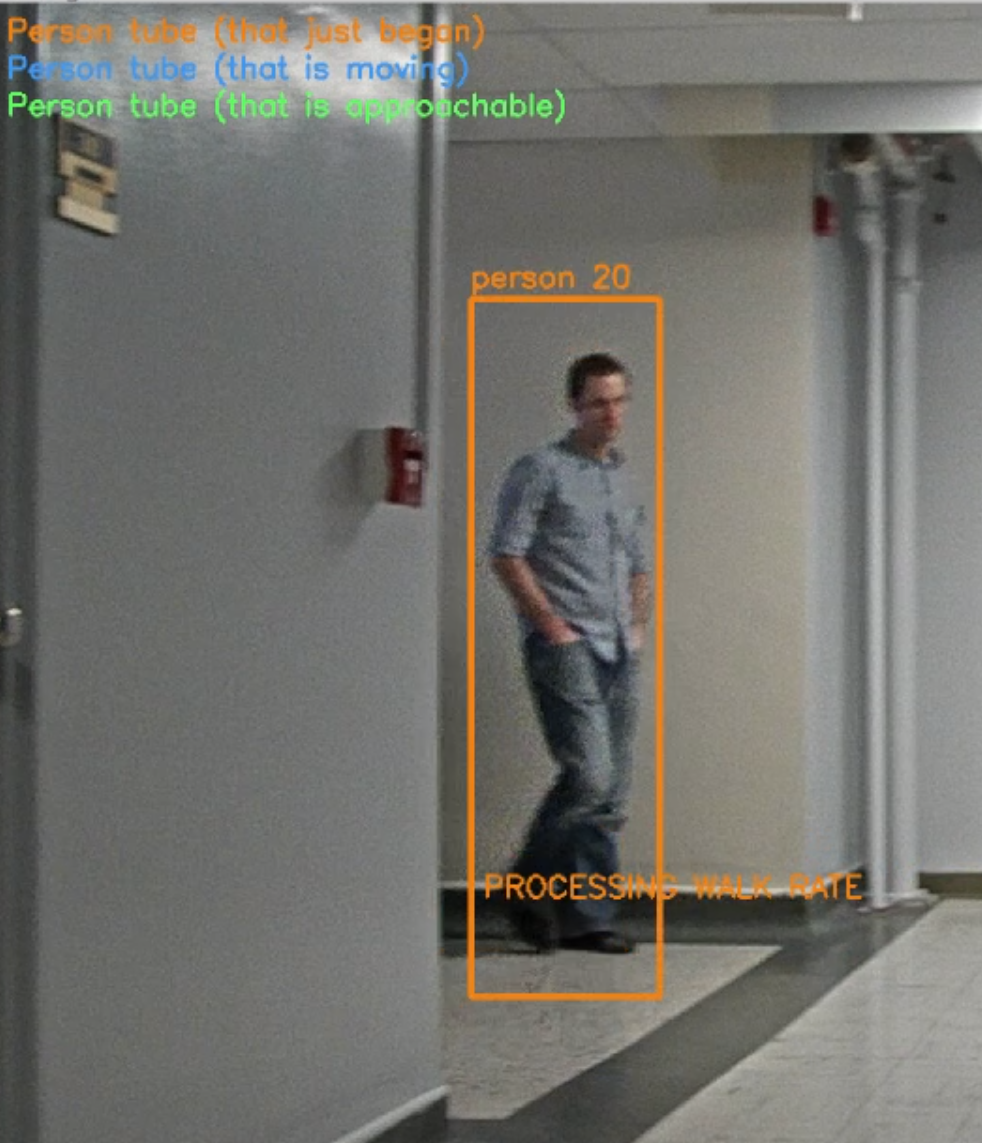}&
    \includegraphics[height=100pt]{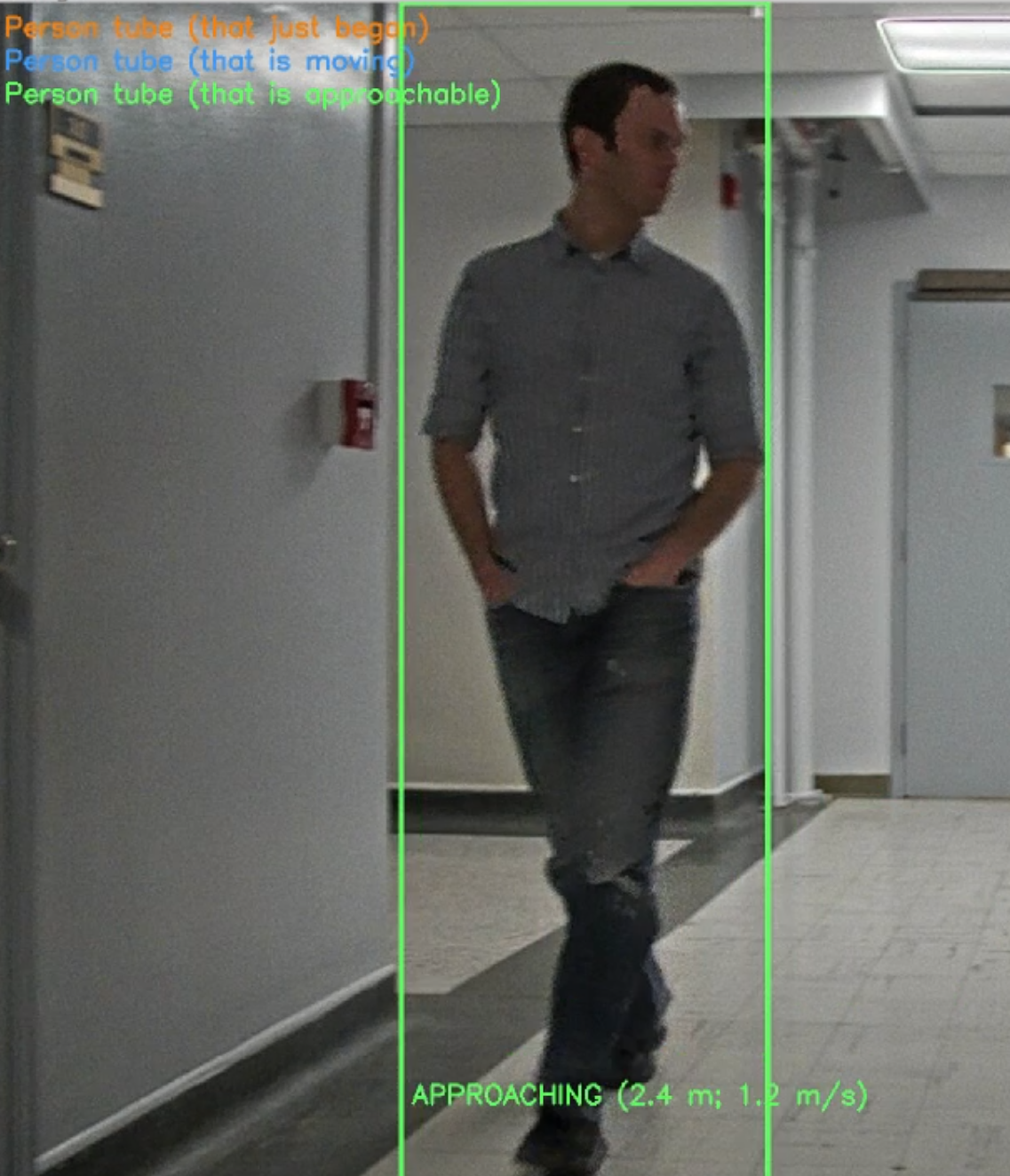}\\
    \includegraphics[height=100pt]{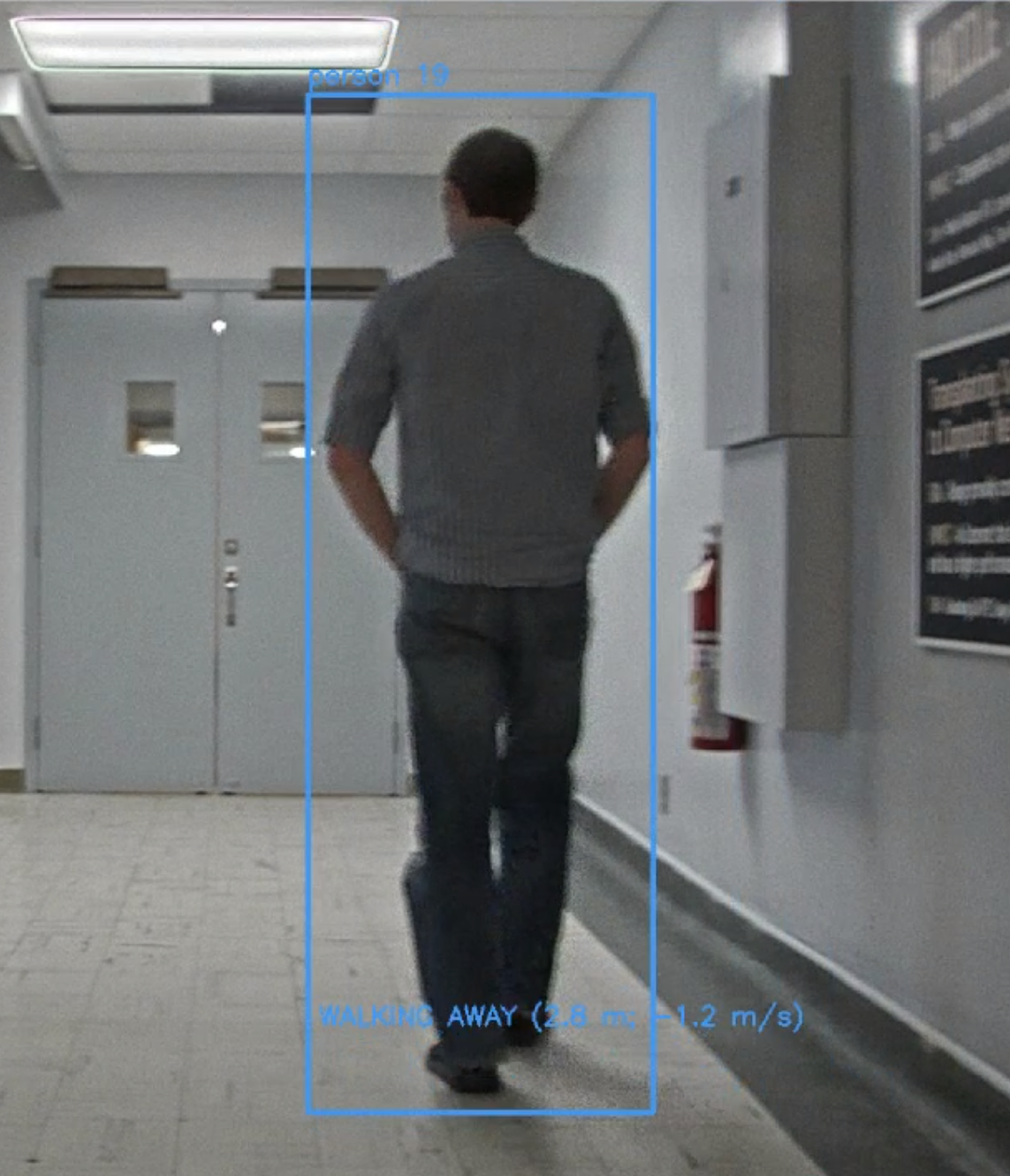}&
    \includegraphics[height=100pt]{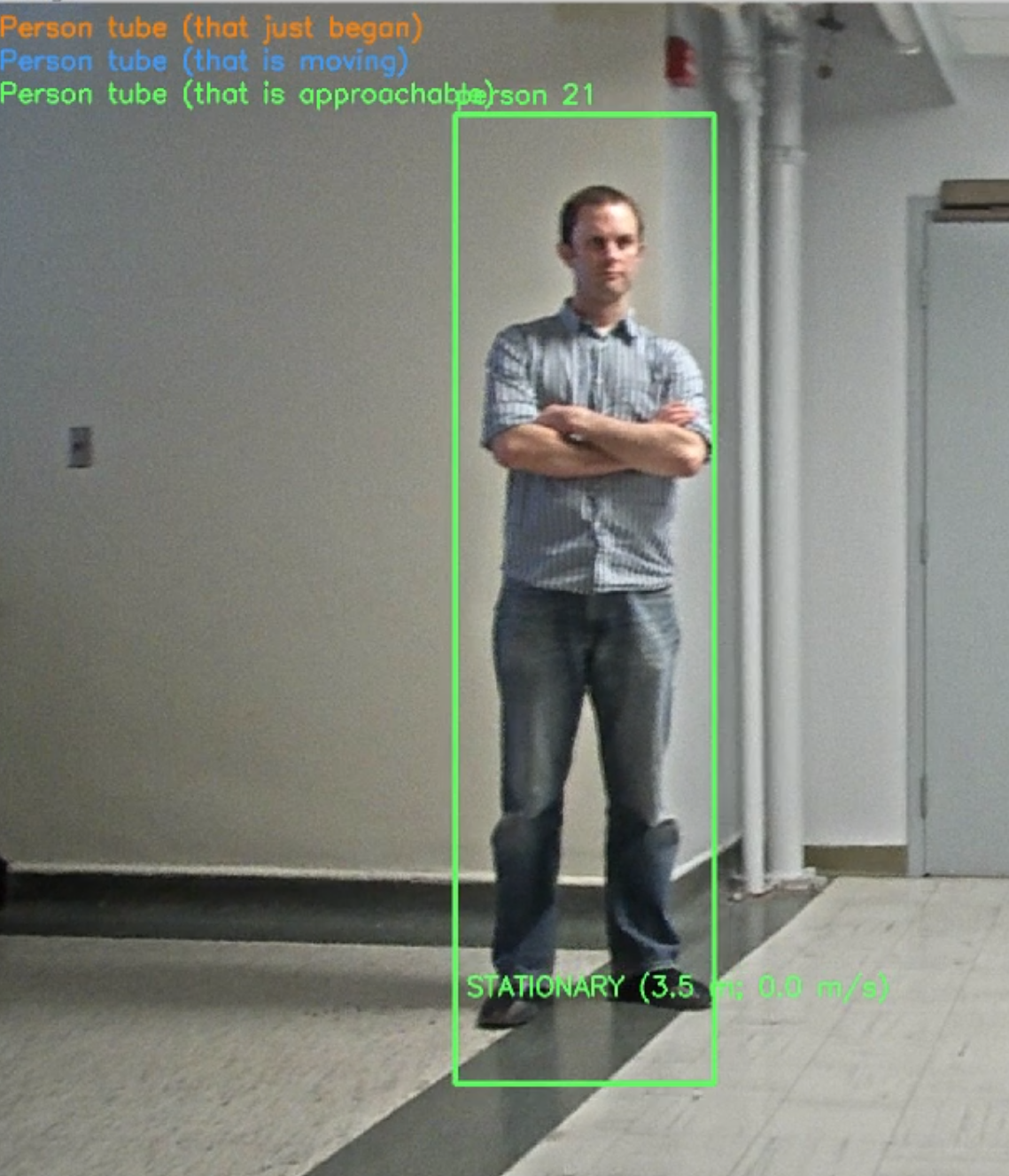}
  \end{tabular}}
  \caption{Approachable person detector output.
    Tracks that have just begun and whose walk rate/polarity are still being
    determined are colored orange.
    Tracks that are walking away are colored blue.
    Tracks that meet the criteria for being either stationary or
    approaching the robot, are colored green.
    Near the bottom of each bounding box, the person's distance from the robot
    is displayed on the left and their walking speed is displayed on the right.
    Positive implies approaching the robot; negative implies
    walking away.}
  \label{fig:approachable-person}
\end{figure}

Tracks are classified based on the length of time they have been active and the
average walking speed of the person.
Once a track is $\geq$1.5~s long, the average walking speed is computed
over the previous 1.5~s.
The person's walking direction is determined by whether the most recent
location is closer to the robot than the location $\approx$1.5~s ago.
If the most recent location is closer by more than 0.3~m/s, the person
is considered \emph{approaching the robot}.
If the most recent location is further by more than 0.3~m/s, the person
is considered \emph{walking away}.
Otherwise, the person is considered \emph{stationary}.
Tracks that are $<$1.5~s long are considered as having just begun and are thus
ignored as part of any decision making on the part of the robot.
Sample outputs of each of these four situations are shown in
\figref{fig:approachable-person}.

Tracks that are classified as stationary or approaching the robot are
considered ``approachable.''
When such occurs, the robot transitions to the \approachperson\ state.

\subsection{\approachperson}

Once an approachable person is successfully located, the system enters the
\approachperson\ state to introduce itself to the person, drive up to them,
and face them just as a human would.
Using the person's location in the quantitative map, the robot computes a
driving goal that is 0.8~m from the person (in a direct path from the robot to
the person) with a pose that is facing the person, and begins driving there.
If the person is approaching the robot, once they are within 7~m, the
robot will hail them and ask them for help.
When the robot is $\leq$2~m from the person, it will introduce itself to
initiate the conversation.

Throughout this process, the person's location in the quantitative map is being
monitored and updated, and new potential driving-goal locations are being
computed.
If any of these are $\geq$0.5~m from the previous driving goal, the robot
updates its driving goal.
This allows the robot to approach the person in a closed-loop fashion.
If instead of stopping, the person chooses to walk up to the robot, once they
are within 0.8~m of the robot, the robot will immediately stop and consider this
as having successfully approached the person.
It will continue the process of monitoring and updating the driving goal until
it either reaches the person, they leave the field of view (\eg\ enter a room),
or begin to walk away.
If either of the latter two scenarios occur, the robot will begin this entire
process over with the next-closest approachable person.
If there are none present, it returns to the \wander\ state.

\subsection{\holdconversation}

Once the robot approaches the person, it enters the \holdconversation\
state whose objective is to construct a plan of how to navigate to a specific
destination by engaging in dialogue with the person.
This plan is represented as a list of navigational actions that can be executed
by the robot as commands.
If the person were to formally communicate the plan to the robot, this task
would be trivial.
What makes this nontrivial is that the communication is done informally through
spoken natural language.

Inferring a plan from spoken dialogue with a person presents several challenges.
First, people often ramble and speak incoherently.
Second, Google's Speech-to-Text API is imperfect and sometimes returns incorrect
and/or nonsensical results.
Third, state-of-the-art parsers, trained on written text rather than spoken
text, create abnormal parse trees that make extracting relevant and accurate
information more difficult.

\begin{table*}
  \centering
  \caption{Various kinds of actions that can fill plan steps as produced by
    \holdconversation.}
  \label{tab:conversation-domain}
    \resizebox{0.8\linewidth}{!}{\begin{tabular}{lll}
      \hline
      Kind & Action & English description \\
      \hline
      $\textit{dir}$
      & \texttt{forward}
      & drive \emph{\textbf{forward}} until a stop condition is encountered \\
      & \texttt{left}
      & rotate 90\degree\ left \\
      & \texttt{right}
      & rotate 90\degree\ right \\
      & \texttt{turn-around}
      & rotate 180\degree \\
      & \texttt{either}
      & \texttt{left} or \texttt{right} depending on availability \\
      \hline
      $\textit{int}$
      & \texttt{elbow}
      & intersection of type \textbf{elbow} \\
      & \texttt{three-way}
      & intersection of type \textbf{three-way} \\
      & \texttt{four-way}
      & intersection of type \textbf{four-way} \\
      & \texttt{int-L}
      & intersection with a \textsf{left} hallway trajectory \\
      & \texttt{int-R}
      & intersection with a \textsf{right} hallway trajectory \\
      & \texttt{int-F}
      & intersection with a \textsf{left} and/or \textsf{right} hallway
        trajectory, as well as a \textsf{forward} hallway trajectory \\
      & \texttt{end}
      & intersection with a \textsf{left} and/or \textsf{right} hallway
        trajectory, but no\textsf{forward} hallway trajectory \\
      \hline
      $\textit{goal}$
      & \texttt{goal-F}
      & goal is somewhere up ahead \\
      & \texttt{goal-L}
      & goal is somewhere up ahead on the left \\
      & \texttt{goal-R}
      & goal is somewhere up ahead on the right \\
      & \texttt{person}
      & our next task is to find a person for further directions \\
      \hline
    \end{tabular}}
\end{table*}

To overcome these obstacles, we employ the following general approach to plan
inference, \ie\ constructing a plan from natural-language dialogue.
We maintain a plan that consists of a sequence of steps, that are filled with
actions of various kinds, including directions, intersections, and goals.
Steps in the plan may be unfilled, denoted by $\Box$.
Initially, the plan consists of a single unfilled step.
The current (partial) plan is used at every dialogue turn to generate a query
to the person.
The response is processed in the context of the current (partial) plan to fill
in unfilled steps, potentially add new steps that are either filled or
unfilled, and potentially change or delete steps.
This processing takes two forms.
One is a sequence of steps for filling in parts of the plan based on the
person's response.
The other is a set of rewrite rules that update parts of the plan based on the
local context in the plan.
If, after processing, the plan contains one or more unfilled steps, a new query
is generated for the person and processing continues.
This process continues until the plan contains no unfilled steps.
To prevent an indefinitely long conversation, the length of the plan is
restricted to ten steps.
If a plan gets to that length, the robot ends the conversation, executes those
steps, then searches for another person to ask.

The sequence of plan actions must be coherent.
The rewrite rules attempt to render plans coherent.
For example, if the plan indicates that the robot eventually reaches an
\textbf{elbow} in the hallway, it would be incoherent for it to drive straight
at the \textbf{elbow}.
The robot uses its current knowledge to formulate and pose a query about a
missing piece of information.
The person's response might provide information about one or more steps,
or change a step that was incorrect.
Rewrite rules are applied to fill in or modify steps that may not have been
explicitly given but were implied.
This process is repeated until a complete plan is constructed.

\subsubsection{Spoken Communication}

Our robot converses in real time, adhering to the norm of taking turns while
talking and using common sense to factor in previous utterances into subsequent
utterances.
Speech recognition and speech synthesis are mutually exclusive operations so
a person's response is only processed when they finish speaking.

\subsubsection{Information Extraction}

In the \holdconversation\ state, the robot attempts to fill in the steps of a
plan with actions from a natural-language utterance.
Our plans are formulated with a certain plan structure that stipulates that
adjacent steps form pairs that constitute commands to the robot.
The first step in a pair must be a direction action.
The second step in a pair will be a stop condition and takes the form of either
an intersection action or a goal action.
Only the last step in the plan can be a goal action.
All other stop conditions must be intersection actions.
Thus a plan can be though of as a sequence of commands, each command being a
direction for the robot to drive in until the specified stop condition, with
the robot ultimately checking that it has reached the goal.
The various actions that can fill plan steps of various kinds are shown in
\tabref{tab:conversation-domain}.

\begin{figure*}
  \centering
  \includegraphics[width=0.9\linewidth]{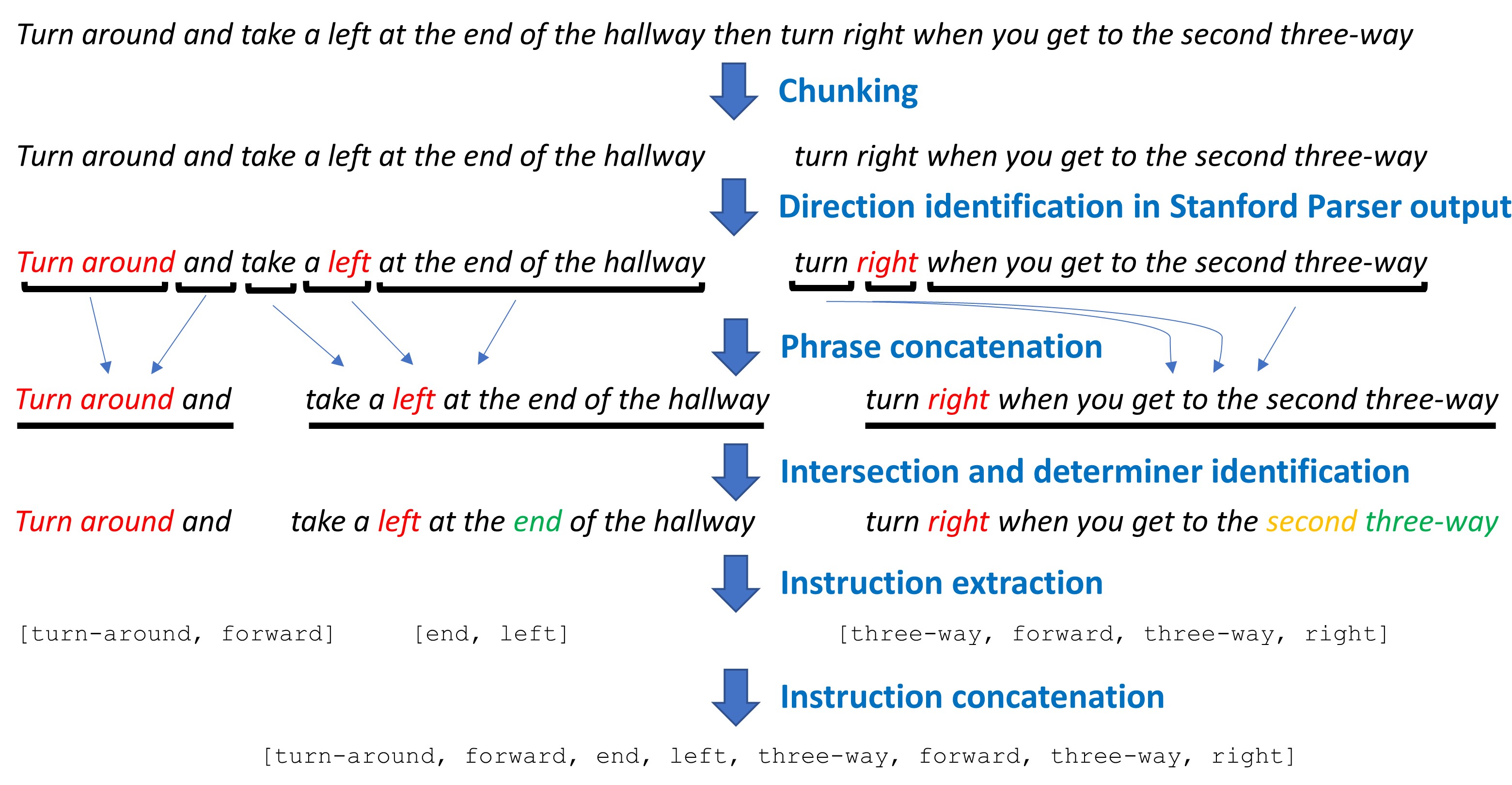}
  \caption{Instruction extraction via parsing steps.}
  \label{fig:conversation-example}
\end{figure*}

Prior to any dialogue, the plan is initialized with a single empty step:
$[\Box]$.
In order to populate the plan, the robot's first question to the person is
`Can you tell me how to navigate to $\langle$destination$\rangle$?'
The steps we use to parse the response and populate the plan with actions are
illustrated in \figref{fig:conversation-example}.
First, the utterance is chunked by splitting it at key elements such as `.'
or `then.'
Each chunk is fed into the Stanford Parser.
This results in more accurate parse trees than feeding the whole utterance into
the Stanford Parser.
Then, the parse trees are searched for phrases that contain direction
keywords (\eg\ `turn around,' `left,' or `right') to insert the
corresponding actions into the plan.
If no direction keywords are found, the utterance is searched for any
intersection or goal keywords to insert the corresponding actions into the
plan.

Each phrase that contains one of the direction keywords is concatenated
with its preceding and succeeding phrase to provide context for information
extraction.
Starting with the leaf node in the parse tree that corresponds to the direction
keyword, the tree is searched for the nearest parent with one of the following
POS tags: ADJP, ADVP, CONJP, FRAG, INTJ, LST, NAC, NP, NX, PP, PRN, PRT, QP,
RRC, UCP, VP, WHADJP, WHAVP, WHNP, or WHPP\@.
The trees corresponding to the left and right siblings of this parent node are
used as the preceding and succeeding phrases, respectively.
The preceding phrase is further refined by searching for the first verb that
precedes the direction keyword; if that does not exist, we search for the first
preposition that precedes the direction keyword.
All words preceding the verb or preposition are eliminated from the preceding
phrase.
If no verb or preposition is found, no preceding phrase is used.
In the event that the direction keyword's parent has no left or right siblings,
we use the entire tree from the chunk as the concatenated phrase.

Each set of concatenated phrases is searched for the presence of an
intersection keyword, direction and intersection determiners (\eg\ `first,'
`second,' or `last'), and destination verbs (\ie\ `be,' `find,' `see').
If a destination verb is present, a goal action is created based on the
direction keyword (\ie\ `left' $\rightarrow$ \texttt{goal-L}, `right'
$\rightarrow$ \texttt{goal-R}, otherwise \texttt{goal-F}).
Otherwise, the direction keyword, direction determiner, intersection keyword,
and intersection determiner are used to populate the plan with corresponding
actions.
For example, in the right chunk in \figref{fig:conversation-example}, it can be
inferred from the determiner `second' for the intersection keyword
`three-way' that the plan should contain two steps, each with a
\texttt{three-way} intersection action.
This also implies that the robot should go straight at the first one and turn
\textsf{right} at the second one.
If there were no direction keywords found, the phrase is searched for
intersection or goal keywords that may have been missed, adding in the
corresponding plan actions.
The plan actions extracted from each chunk are concatenated to create the final
plan.
For a given utterance, this plan must be consistent with plan structure as
described above to ensure proper execution.
\begin{compactenum}[1.]
\item The first step must be a direction action.
\item The last step must be a goal action.
\item A direction action must always follow an intersection action.
\item Two intersection actions cannot be adjacent.
\item Two direction actions cannot be adjacent (unless the first one is
  turn-around).
\end{compactenum}
These properties of plan structure allow us to formulate patterns to
describe all minimal invalid action sequences.
Each such invalid action sequence can be rendered valid by appropriately
adding, deleting, or modifying plan steps.
Such plan modification is performed by rewrite rules.
For example, consider the plan $[\texttt{forward}, \texttt{left}]$, which
implies that the robot must arrive at an intersection with a \textsf{left} turn.
This plan can be rewritten as $[\texttt{forward}, \Box, \texttt{left}]$ to
indicate that it needs to populate a plan step with an action that specifies
the intersection type at which it will make the \textsf{left} turn.
These rewrite rules are repeatedly applied to the plan until it no longer
contains any invalid action sequences.
\tabref{tab:rewrite-rules} shows the complete list of plan rewrite rules.

\begin{table}
  \centering
  \caption{Plan rewrite rules.
    In the following, $\textit{dir}$ denotes any direction action,
    $\textit{ntadir}$ denotes any direction action except \texttt{turn-around},
    $\textit{int}$ denotes any intersection action, and
    $\textit{goal}$ denotes any goal action.
    The first matching rule applies when multiple rules match.}
  \label{tab:rewrite-rules}
  \begin{equation*}
    \begin{array}{lcl}
      \hline
      \textit{goal}&\leadsto&
      \Box\;\textit{goal}\\
      \ldots\;\textit{goal}\;\Box&\leadsto&
      \ldots\;\textit{goal}\\
      \Box\;\texttt{int-L}\;\textit{dir}\;\ldots&\leadsto&
      \texttt{forward}\;\texttt{int-L}\;\textit{dir}\;\ldots\\
      \Box\;\texttt{int-R}\;\textit{dir}\;\ldots&\leadsto&
      \texttt{forward}\;\texttt{int-R}\;\textit{dir}\;\ldots\\
      \textit{int}\;\texttt{turn-around}\;\ldots&\leadsto&
      \texttt{turn-around}\;\ldots\\
      \texttt{turn-around}\;\textit{int}\;\ldots&\leadsto&
      \texttt{turn-around}\;\texttt{forward}\;\textit{int}\;\ldots\\
      \textit{int}\;\ldots&\leadsto&
      \Box\;\textit{int}\;\ldots\\
      \ldots\;\texttt{forward}\;\texttt{forward}\;\ldots&\leadsto&
      \ldots\;\texttt{forward}\;\ldots\\
      \ldots\;\textit{goal}\;\ldots&\leadsto&
      \ldots\;\ldots\\
      \ldots\;\texttt{elbow}\;\texttt{int-L}\;\ldots&\leadsto&
      \ldots\;\texttt{elbow}\;\ldots\\
      \ldots\;\texttt{elbow}\;\texttt{int-R}\;\ldots&\leadsto&
      \ldots\;\texttt{elbow}\;\ldots\\
      \ldots\;\texttt{three-way}\;\texttt{int-L}\;\ldots&\leadsto&
      \ldots\;\texttt{three-way}\;\ldots\\
      \ldots\;\texttt{three-way}\;\texttt{int-R}\;\ldots&\leadsto&
      \ldots\;\texttt{three-way}\;\ldots\\
      \ldots\;\texttt{four-way}\;\texttt{int-L}\ldots&\leadsto&
      \ldots\;\texttt{four-way}\;\ldots\\
      \ldots\;\texttt{four-way}\;\texttt{int-R}\ldots&\leadsto&
      \ldots\;\texttt{four-way}\;\ldots\\
      \ldots\;\textit{ntadir}\;\texttt{forward}\;\ldots&\leadsto&
      \ldots\;\textit{ntadir}\;\ldots\\
      \ldots\;\textit{ntadir}\;\textit{dir}\;\ldots&\leadsto&
      \ldots\;\textit{ntadir}\;\Box\;\textit{dir}\;\ldots\\
      \ldots\;\textit{int}_1\;\textit{int}_2\;\ldots&\leadsto&
      \ldots\;\textit{int}_1\;\Box\;\textit{int}_2\;\ldots\\
      \ldots\;\textit{int}\;\textit{goal}\;\ldots&\leadsto&
      \ldots\;\textit{int}\;\Box\;\textit{goal}\;\ldots\\
      \ldots\;\textit{dir}&\leadsto&
      \ldots\;\textit{dir}\;\Box\\
      \ldots\;\textit{int}&\leadsto&
      \ldots\;\textit{int}\;\Box\\
      \ldots\;\texttt{elbow}\;\texttt{forward}\;\ldots&\leadsto&
      \ldots\;\texttt{elbow}\;\ldots\\
      \ldots\;\texttt{elbow}\;\Box\;\ldots&\leadsto&
      \ldots\;\texttt{elbow}\;\texttt{either}\;\ldots\\
      \Box\;\texttt{forward}\;\ldots&\leadsto&
      \texttt{forward}\;\ldots\\
      \Box\;\texttt{turn-around}\;\ldots&\leadsto&
      \texttt{turn-around}\;\ldots\\
      \hline
    \end{array}
  \end{equation*}
  \begin{tabular}{lll}
  \end{tabular}
\end{table}

\subsubsection{Dialogue}

The goal of \holdconversation\ is to construct a complete plan, with no
unfilled steps.
If the plan is not complete, a query is generated and posed to the person for
the first unfilled step in the plan.
Information is extracted from the person's response to fill the
corresponding unfilled step in the plan.
The type of the query depends on the pattern in which the unfilled step occurs.
There are two query types: \emph{single}, which requests a single piece of
information, and \emph{open-ended}, which gives the person more freedom in
their response.
\Tabref{tab:query-templates} illustrates the different queries for each pattern
and query-type.

If the query-type is single, \holdconversation\ looks for the
presence of that particular type of information in the response.
After finding it, all words up to it are removed and the remaining text is
parsed in an open-ended fashion.
Consider an example:
\begin{quote}
  \begin{tabular}{lp{150pt}}
    Robot: & Which direction do I start out going?\\
    Person: & you start left, then turn right at the end of the hallway
    \end{tabular}
\end{quote}
Our parsing method extracts `left' as the answer to the query and then
replaces the corresponding $\Box$ in the plan with \texttt{left}.
Then it parses `then turn right at the end of the hallway' in an
open-ended manner and inserts the extracted instructions into the plan
after \texttt{left}.
If the query-type is open-ended, the response is parsed in the same way
as the response to the robot's first question.
The robot repeatedly poses questions, processes the respond, and applies
rewrite rules until it generates a complete plan with no missing information.
A complete and consistent plan indicates success and results in a transition to
the \followdirections\ state.

If the plan is complete but not consistent, or if no plan could be constructed,
then the plan $[\texttt{right}, \texttt{person}]$ is created and a transition to
the \followdirections\ state is made.
This short plan causes the robot to rotate away from the person so that they are
no longer in the field of view and will not be redetected as an approachable
person.
After rotating 90\degree\ to the right, the robot transitions to the \wander\
state and will seek out a new person to ask for directions.

\subsubsection{Addressing Corner Cases}

To help facilitate a more natural conversation, we have a small amount of code
to address a few corner cases that may arise in spoken conversation.
If a response to a query is not heard within 5~s, the robot states this fact
and repeats its query.
If the robot is not able to extract any useful information from a response, it
indicates such and may provide some information about what it does understand
(\eg\ directions and intersections).
If the robot goes two turns without the plan changing (\eg\ it fails to
understand the person's instructions or it hears no response), the robot ends
the conversation and carries out whatever portion of the plan is usable.
If the person indicated the robot misunderstood their last utterance, the robot
backs up one step, using the previous iteration of the plan and its
corresponding query.
If the person indicates that they would like to start over, the robot resets
the plan and asks its original query.
Some sample conversations from our trials are shown in
\tabref{tab:sample-convs}.

\begin{table}
  \centering
  \caption{Query templates for plan patterns.
    The notation $\langle$nth$\rangle$ refers to a direction determiner
    generated based on how many direction actions of the same type appear
    prior to $\textit{dir}$ in the current partial plan.
    The notation $\langle$jth$\rangle$ refers to an intersection determiner
    generated based on how many intersection actions of the same type appear
    prior to $\textit{int}$ in the current partial plan.
    The first matching template applies when multiple templates match.}
  \label{tab:query-templates}
  \resizebox{\linewidth}{!}{\begin{tabular}{llp{140pt}}
    \hline
    Pattern
    & Query-type
    & Query \\
    \hline
    $\Box$
    & open-ended
    & `Could you tell me how to navigate to $\langle$destination$\rangle$?' \\
    $\Box\;\textit{int}\;\ldots$
    & single
    & `Which direction do I start out going?' \\
    $\Box\;\textit{goal}$
    & single
    & `Which direction do I start out going?' \\
    $\texttt{turn-around}\;\Box\;\ldots$
    & open-ended
    & `What do I do after turning around?' \\
    $\textit{dir}\;\Box\;\ldots$
    & open-ended
    & `What do I do after starting to go $\textit{dir}$?' \\
    $\texttt{turn-around}\;\textit{dir}\;\Box\;\ldots$
    & open-ended
    & `What do I do after turning around and going $\textit{dir}$?' \\
    $\ldots\;\texttt{left}\;\ldots\;\texttt{left}\;\Box\;\ldots$
    & open-ended
    & `What do I do after I turn left (this being the $\langle$nth$\rangle$
       left I take)?' \\
    $\ldots\;\texttt{left}\;\Box\;\ldots$
    & open-ended
    & `What do I do after I turn left?' \\
    $\ldots\;\texttt{right}\;\ldots\;\texttt{right}\;\Box\;\ldots$
    & open-ended
    & `What do I do after I turn right (this being the $\langle$nth$\rangle$
       right I take)?' \\
    $\ldots\;\texttt{right}\;\Box\;\ldots$
    & open-ended
    & `What do I do after I turn right?' \\
    $\ldots\;\texttt{elbow}\;\textit{dir}\;\Box\;\ldots$
    & open-ended
    & `Where do I go after the elbow (this being the $\langle$jth$\rangle$
       elbow)?' \\
    $\ldots\;\textit{int}\;\texttt{forward}\;\Box\;\ldots$
    & open-ended
    & `What do I do after going through the $\langle$jth$\rangle$
       $\textit{int}$?' \\
    $\ldots\;\textit{int}\;\Box$
    & open-ended
    & `What do I do after getting to the $\langle$jth$\rangle$
      $\textit{int}$?' \\
    $\ldots\;\textit{int}_1\;\Box\;\textit{int}_2\;\ldots$
    & single
    & `When I am at $\textit{int}_1$, which direction
      will I go to get to $\textit{int}_2$?' \\
    \hline
  \end{tabular}}
\end{table}

\begin{table}
  \centering
  \caption{Sample conversations from our trials.}
  \label{tab:sample-convs}
  \resizebox{\linewidth}{!}{\begin{tabular}{lp{240pt}}
    \hline
    Speaker & Utterance \\
    \hline
    Robot: & Could you tell me how to navigate to 345? \\
    Person: & yeah, turn around go to the end of the hall and you'll
              take a lot to the bathroom. \\
    Robot: & What do I do after getting to the first end of the hall? \\
    Person: & you take a left at the bath. \\
    Robot: & What do I do after I turn left? \\
    Person: & app \\
    Robot: & I did not understand that.
             What do I do after I turn left? \\
    Person: & you go to the end of the hall. \\
    Robot: & What do I do after getting to the second end of the hall? \\
    Person: & turn right. \\
    Robot: & What do I do after I turn right? \\
    Person: & it'll be the third door on the left. \\
    Robot: & Thanks for your help. Have a great day! \\
    \hline
    Plan: & [\texttt{turn-around}, \texttt{forward}, \texttt{end},
             \texttt{left}, \texttt{end}, \texttt{right},
             \texttt{goal-L}] \\*[2ex]
    Robot: & Could you tell me how to navigate to 276? \\
    Person: & yeah, turn around then turn right then your first left and then
              the door will be on your left. \\
    Robot: & Thanks for your help. Have a great day! \\
    \hline
    Plan: & [\texttt{turn-around}, \texttt{forward}, \texttt{int-R},
             \texttt{right}, \texttt{int-L}, \texttt{left},
             \texttt{goal-L}] \\*[2ex]
    Robot: & Could you tell me how to navigate to 1273? \\
    Person: & yes, turn right. \\
    Robot: & What do I do after I turn right? \\
    Person: & \\
    Robot: & I did not hear you.
             What do I do after I turn right? \\
    Person: & and then turn right. \\
    Robot: & What do I do after I turn right (this being the second right I
             take)? \\
    Person: & find room 1273. \\
    Robot: & Thanks for your help. Have a great day! \\
    \hline
    Plan: & [\texttt{forward}, \texttt{int-R}, \texttt{right}, \texttt{int-R},
             \texttt{right}, \texttt{goal-F}] \\
    \hline
  \end{tabular}}
\end{table}

\subsection{\followdirections}

With a successfully extracted plan, the robot enters the \followdirections\
state.
The plan includes direction, intersection, and goal actions.
Direction and intersection actions are grounded in the environment by the
navigation process described in \secref{sec:navigation-process}.
The goal action is a transition condition to \followdirections\ that it has
completed the plan provided by the person and should transition to the next
state.

\subsubsection{Plan Preprocessing}

In \secref{sec:wander-fsm}, we described the \drivethroughintersection\ substate
and its purpose in ensuring the robot exited the registered intersection before
being able to detect another registered intersection and re-evaluate what to do.
That behavior is important to \followdirections\ as well for a similar reason:
if the robot entered a registered intersection, rotated in the direction
specified in the plan, and immediately began looking for the subsequent
intersection, it could mistakenly think it had reached it, despite still being
in the same registered intersection.
We want the robot to completely exit the current registered intersection before
beginning to look for the subsequent one.
To facilitate this behavior, \followdirections\ takes the plan received
from \holdconversation, searches for all but the last instance of the pattern
$[\textit{int}, \textit{dir}]$, and inserts the following two
actions after the direction action:
$[\texttt{forward-through-int}, \texttt{forward}]$.
As an example, the plan
$[\texttt{forward}, \texttt{elbow}, \texttt{left}, \texttt{elbow},
  \texttt{left}, \texttt{goal-F}]$
would become

$\begin{array}{l}
  [\texttt{forward}, \texttt{elbow}, \texttt{left},
   \texttt{forward-through-int},\\
   \;\texttt{forward}, \texttt{elbow},
   \texttt{left}, \texttt{goal-F}].
 \end{array}$

\subsubsection{\followdirections\ Substates}
\label{sec:follow-directions-fsm}

The \followdirections\ state has five substates:
\begin{inparaenum}
\item \makedecision,
\item \driveforward,
\item \rotate,
\item \drivethroughintersection, and
\item \complete.
\end{inparaenum}
\followdirections\ maintains a step counter, indicating the current step to
execute.
It enters the \makedecision\ substate first, initializing the step counter to
the first step in the plan.
When the current action is \texttt{forward}, it will enter the
\driveforward\ substate wherein it drives \emph{\textbf{forward}} until the
subsequent intersection in the plan is found.
In the example $[\texttt{forward}, \texttt{elbow}, \texttt{left}, \ldots]$
the robot would drive \emph{\textbf{forward}} until it detects an \texttt{elbow}
intersection.
As noted in \tabref{tab:conversation-domain}, some of the intersection keywords
are less specific and only contain an indication of what hallway trajectories
one would expect to find at the specified intersection.
For example, \texttt{int-L} would specify a registered intersection of any type
that includes a hallway trajectory labeled with the qualitative direction
\textsf{left}.

Once the specified intersection has been detected, the robot will return to
the \makedecision\ substate to determine what it needs to do next.
It the next step in the plan contains a direction action that requires
rotation (\ie\ \texttt{left}, \texttt{right}, or \texttt{turn-around}), the
robot will enter the \rotate\ substate.
In this substate, the robot will rotate in-place by a specified amount according
to the direction specified in the plan:
\texttt{left} $\rightarrow$ 90\degree,
\texttt{right} $\rightarrow$ $-$90\degree, and
\texttt{turn-around} $\rightarrow$ 180\degree.
After rotating, the robot will return to the \makedecision\ substate.
If the subsequent step is \texttt{forward-through-int}, the robot enters the
\drivethroughintersection\ substate.
As before, it requires that the intersection type change and the robot travel
at least 2~m from where it rotated before it exits this substate and returns to
the \makedecision\ substate.

Before each of the substates \driveforward, \rotate, and
\drivethroughintersection\ finish performing their task and return to the
\makedecision\ substate, they increment the step counter accordingly.
Eventually, the step counter will reach the goal action.
When this occurs, \makedecision\ transitions to the \complete\ substate.
This is indicative of \followdirections\ having successfully executed the
plan and the robot will transition to the next state.
When the goal action is \texttt{goal-F}, \texttt{goal-L}, or \texttt{goal-R},
the robot has reached the same hallway as the goal and must look for it.
Thus the robot will transition to the \navigatedoor\ state.
When the goal action is \texttt{person}, the robot has carried out as many
steps as the previous person was able to provide and must now seek out a
new person to ask for instructions.
Thus the robot will transition to the \wander\ state.
A diagram of this FSM is shown in \figref{fig:follow-directions-fsm}.

\subsubsection{Forward Driving Goals}
\label{sec:forward-goals2}

\begin{figure}
  \centering
  \includegraphics[width=0.5\linewidth]{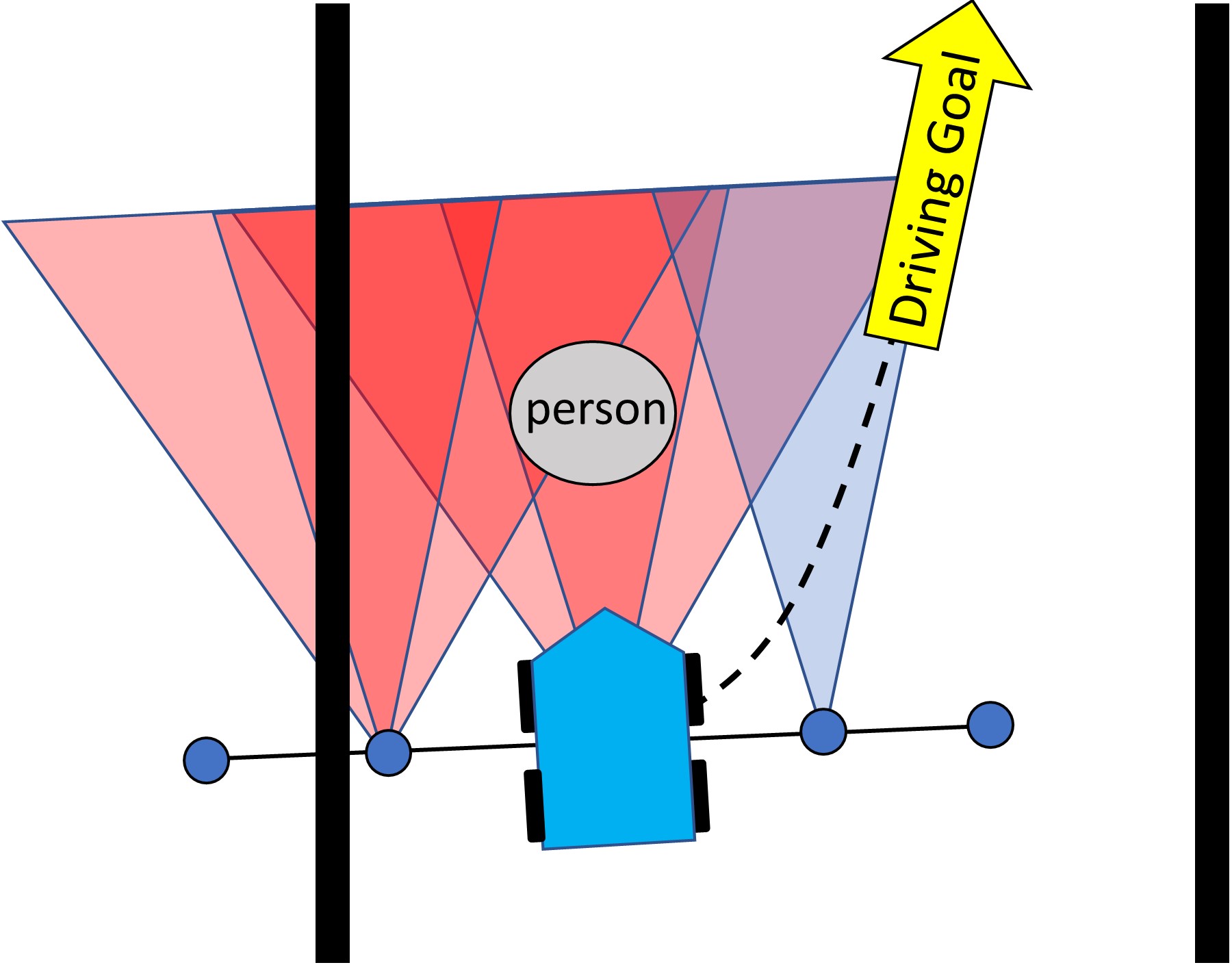}
  \caption{Example of how the robot determines a \emph{\textbf{forward}}
    driving goal when the person it just conversed with has materialized as an
    obstacle in the quantitative map.
    The two transparent red cones coming from the robot indicate that no
    drivable trajectories are free at a cone angle of either
    $\pm\textrm{15\degree}$ or $\pm\textrm{30\degree}$.
    (To prevent this figure from becoming too cluttered, we ignore the cone
    angle of $\pm\textrm{45\degree}$).
    The two transparent red cones coming from the point that is 0.5 meters to
    the \textsf{left} of the robot indicate that there are no drivable
    trajectories at either $\pm\textrm{15\degree}$ or $\pm\textrm{30\degree}$.
    The blue cone coming from the point that is 0.5 meters to the
    \textsf{right} of the robot indicates that a drivable trajectory was found
    and it is used as the \emph{\textbf{forward}} driving goal.}
  \label{fig:forward-goal-person}
\end{figure}

When approaching a person, the robot may have moved to one side of the hallway
and be in a suboptimal pose (similar to that described in
\secref{sec:forward-goals}).
Additionally, because the conversation will have taken some amount of time, the
person may have materialized as an obstacle in the quantitative map,
directly in front of the robot.
If the first action in the plan is \texttt{forward}, the robot may be unable
to do so, even when using the technique described in \secref{sec:forward-goals},
where we search with increasingly wider cone angles in the hopes of finding a
median drivable trajectory.
If no drivable trajectories are available from the robot's pose (which is often
the case because of the robot's proximity to the person), we repeat the same
search for a median drivable trajectory, but use positions that are horizontal
to the robot's pose.
Starting at the robot's current pose, we incrementally search \textsf{left}
then \textsf{right} at distances of $\pm\textrm{0.5~m}$, $\pm\textrm{1~m}$, and
$\pm\textrm{1.5~m}$, with the expectation that at some horizontal position, we
will be far enough away from the person to find a median drivable trajectory.
We then use this median drivable trajectory as our \emph{\textbf{forward}}
driving goal.
This helps to move us around the person and drive in the direction indicated.
Note that this mechanism, also present in the \wander\ state, is the one we
alluded to in \secref{sec:forward-goals}.
\Figref{fig:forward-goal-person} provides a visual explanation of this process.

While in the \driveforward\ substate, the robot will continue driving
\emph{\textbf{forward}} until it either detects the intersection specified in
the plan or reaches the end of the hallway.
If it reaches the end of the hallway without detecting the desired intersection,
the robot will perform a single 360\degree\ spin to ensure that it truly is at a
dead-end.
If, after performing this spin, it has not detected the specified intersection,
it indicates failure to carry out the plan and will return to the
\wander\ state.

\begin{figure*}
  \centering
  \begin{tabular}{@{}c@{}c@{}c@{}}
    \begin{subfigure}[t]{0.33\linewidth}
      \includegraphics[width=\textwidth]{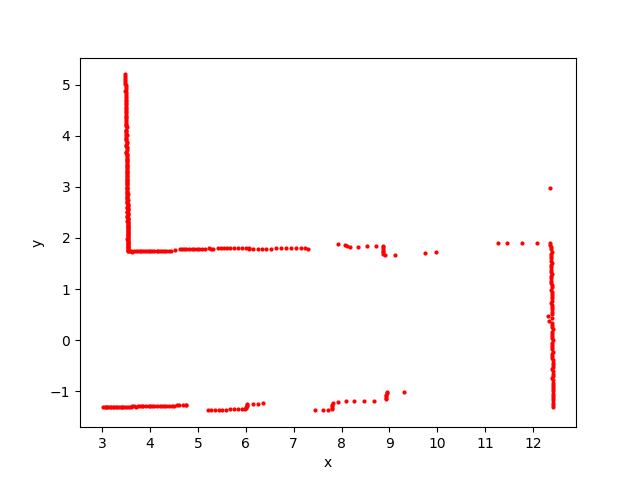}
      \caption{Raw LiDAR points.}
      \label{fig:1}
    \end{subfigure}&
    \begin{subfigure}[t]{0.33\linewidth}
      \includegraphics[width=\textwidth]{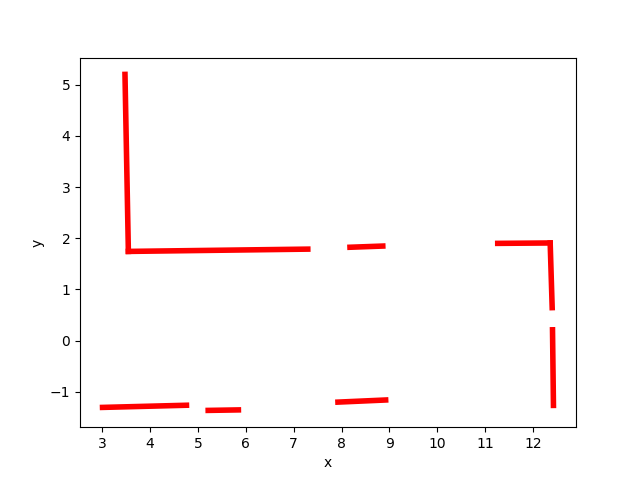}
      \caption{Extracted line segments.}
      \label{fig:2}
    \end{subfigure}&
    \begin{subfigure}[t]{0.33\linewidth}
      \includegraphics[width=\textwidth]{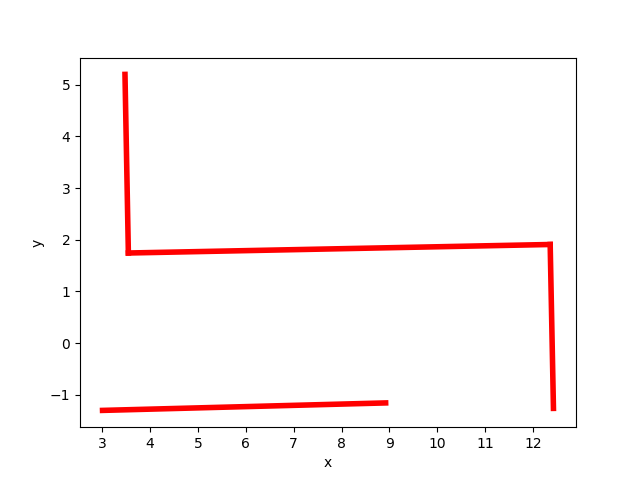}
      \caption{Merged line segments.}
      \label{fig:3}
    \end{subfigure}
  \end{tabular}
  \caption{Process for extracting walls from LiDAR data.}
  \label{fig:segs}
\end{figure*}

\subsection{\navigatedoor}

Once the robot completes execution of the plan with a goal other than
\texttt{person}, it will be located in the hallway containing the desired door
and transitions to the \navigatedoor\ state to conduct a systematic search to
find the door.
The search procedure depends on accurate door localization and common-sense
reasoning.
We leverage several key characteristics of our environment: doors are all of a
similar shape, each door has a door tag displaying the room number, and room
numbers are consecutive odd on one side of the hallway and consecutive even on
the other.
This lets the robot inspect the doors in a logical fashion.

To detect doors, doorways, and elevators, we rely on four standard pieces of
sensor information from the robot:
\begin{inparaenum}
    \item the camera image,
    \item the raw LiDAR data,
    \item the quantitative map, and
    \item the robot pose.
\end{inparaenum}
We use the LiDAR data to first determine regions in the image that can contain
doors, then search these regions for door proposals, and ultimately assign a
confidence score to each proposal.
Because the LiDAR data provides information about obstacle distance
relative to the robot, we can use the robot's position in the quantitative map
to compute absolute locations for the door proposals.

\subsubsection{Detecting Walls}

We rely on 3D information from the LiDAR data to determine the walls.
This data is represented as a collection of 2D $(x,y)$ coordinates that
represent the location of obstacles in the environment surrounding the
robot.
An illustration of such data can be seen in \figref{fig:1}.
We apply the Douglas-Peucker algorithm \citep{DOUGLAS1973ALGORITHMSCARICATURE}
to reduce the number of these points and establish line segments that can
correspond to walls in the environment.
We define each line segment as the 4-tuple $(x_{1}, y_{1}, x_{2}, y_{2})$ where
$(x_{1}, y_{1})$ and $(x_{2}, y_{2})$ correspond to a pair of endpoints.
The resulting line segments can be seen in \Figref{fig:2}.
Due to occlusion or recession, this algorithm can produce disjoint line
segments that correspond to the same wall.
We employ two stages of clustering to merge these disjoint line segments into
\emph{walls}.

Because line segments that correspond to the same wall must share the same
orientation, we cluster the line segments with hierarchical clustering based on
their angles.
We refer to these clusters as \emph{orientation clusters}.
To disambiguate line segments belonging to parallel but distinct walls, we
perform a second stage of clustering.
For each line segment within an orientation cluster, we rotate it by the
negative of its angle so that the resulting line segment is parallel to the
$x$-axis in a standard Cartesian coordinate system.
The transformation is shown in \equref{equation:rotate}.
\begin{equation}
    \label{equation:rotate}
    \begin{bmatrix}
    x'_{1} & x'_{2} \\
    y'_{1} & y'_{2}
    \end{bmatrix} =
    \begin{bmatrix}
    \cos{-\theta} & -\sin{-\theta} \\
    \sin{-\theta} & \cos{-\theta}
    \end{bmatrix}
    \begin{bmatrix}
    x_{1} & x_{2} \\
    y_{1} & y_{2}
    \end{bmatrix}
\end{equation}
Now, for each resulting line segment $(x'_{1}, y'_{1}, x'_{2}, y'_{2})$, its
distance to the $x$-axis is equal to $y'_{1}(=y'_{2})$.
Thus, we cluster all of these line segments based on this distance to
distinguish between parallel walls.
At the end of this step, our final clusters consist of line segments that belong
to the same wall.
For each of these clusters, we merge all of the line segments into a single
line segment $(\hat{x}_{1}, \hat{y}_{1}, \hat{x}_{2}, \hat{y}_{2})$ by taking
the two most extreme endpoints of the line segments.
Each of these merged line segments represents a wall in the robot's immediate
environment.
The final result can be seen in \figref{fig:3}.

\begin{figure}
  \centering
  \includegraphics[width=\linewidth]{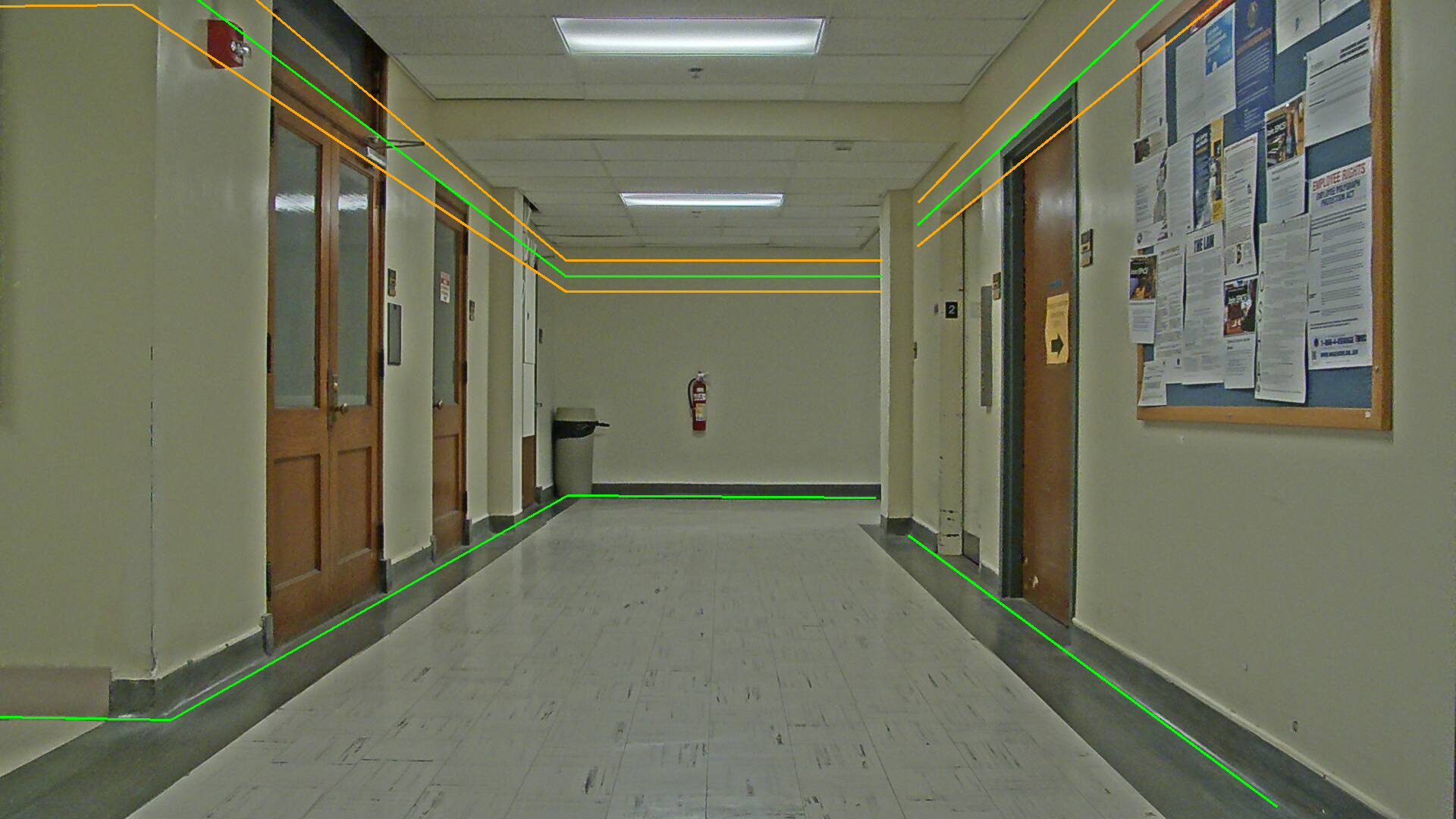}
  \caption{Projected walls with region boundaries.
    Green lines are the tops and bottom of walls projected at heights of 0~m and
    2.2~m respectively, and the orange lines are the $\pm\textrm{15~cm}$
    tolerances for the top.}
  \label{fig:projection}
\end{figure}

Now that we have the locations of the walls, we can create regions in the image
that could contain doors.
Because most doors, doorways, and elevators have a standard height of 2.2~m,
we desire regions that are bounded above by this height and below by the ground
level, 0~m.
Knowing the camera intrinsics, $K$, and extrinsics, $R$ and $t$, we can apply
the transformation in \equref{equation:calib} to project the wall line segments
onto the image at these height levels by setting $h$ appropriately.
\begin{equation}
    \label{equation:calib}
    \begin{bmatrix}
    u_{1} & u_{2} \\
    v_{1} & v_{2} \\
    1     & 1
    \end{bmatrix} =
    K
    \begin{bmatrix}
    R & t
    \end{bmatrix}
    \begin{bmatrix}
    \hat{x}_{1} & \hat{x}_{2} \\
    \hat{y}_{1} & \hat{y}_{2} \\
    h & h \\
    1 & 1
    \end{bmatrix}
\end{equation}
Because we take real data from a moving robot as input, this projection might
not always result in accurate pixel coordinates, which can cause complications
when we search for edges that correspond to the tops of doors.
Therefore, we incorporate a vertical distance tolerance of $\pm\textrm{15~cm}$
for the top boundary.
A visualization of the boundaries and their tolerances can be seen in
\figref{fig:projection}.
Each pair of a top and bottom boundary is considered as a distinct wall region.

\subsubsection{Generating Door Proposals}

\begin{figure}
  \centering
  \includegraphics[width=\linewidth]{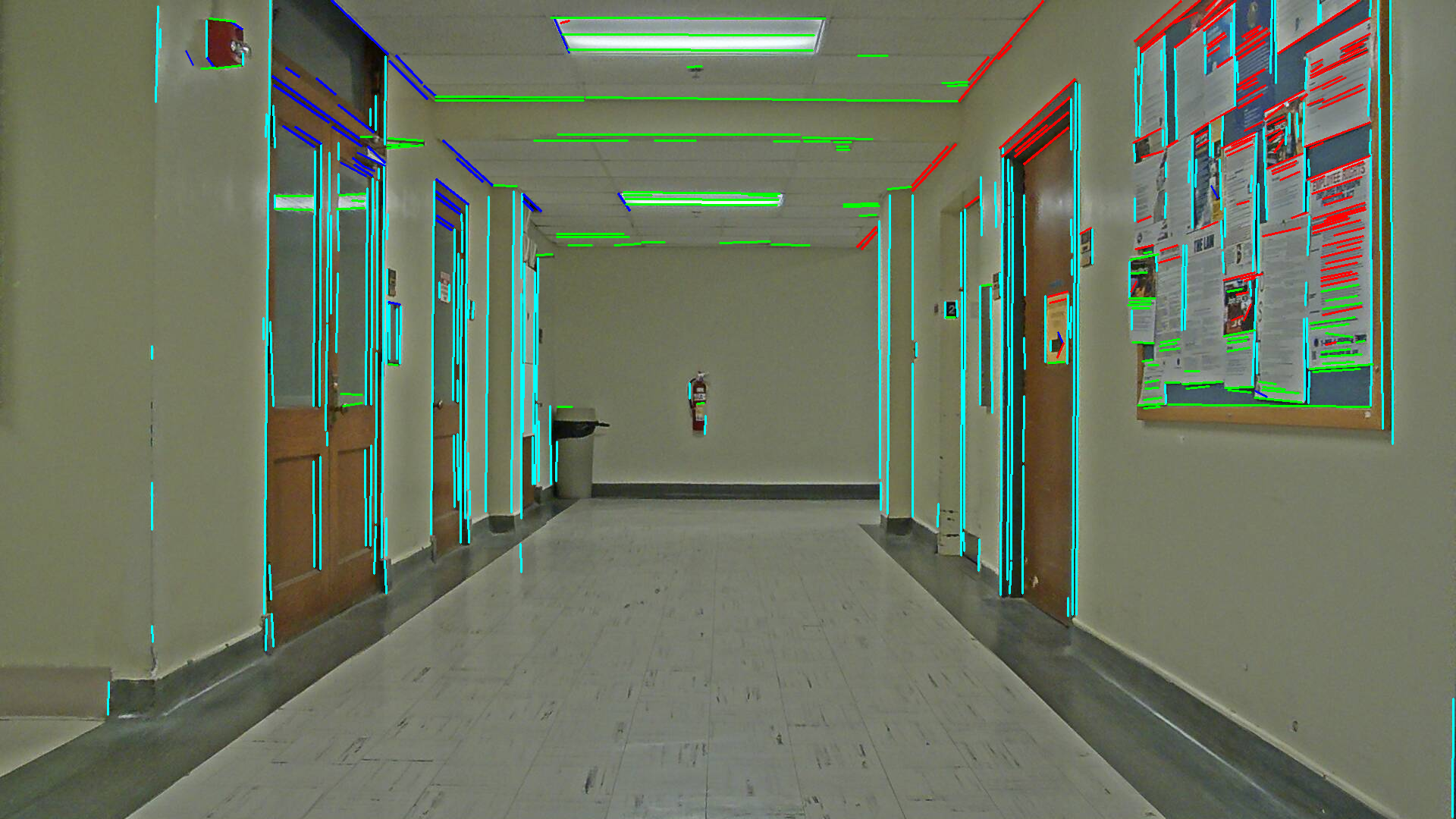}
  \caption{Line segment detections, color coded by their orientation.
    Green lines are horizontal.
    Vertical lines are teal.
    Blue lines have a downward slope.
    Red lines have an upward slope.}
  \label{fig:edges}
\end{figure}

We use edge detection as a basis for generating door proposals, incorporating
mechanisms that are robust to noisy edge detections which are prevalent in
images obtained from a moving robot.
We employ LSD (Line Segment Detector) to detect line segments in the image
without any parameter tuning.
Each line segment is represented as the 4-tuple $(u_{1}, v_{1}, u_{2}, v_{2})$.
Inspired by \citet{Shi2006InvestigatingEnvironments}, we quantize the line
segments into several bins depending on their orientation.
We do this to isolate the line segments that can potentially belong to a door,
specifically the two posts and the top.
Therefore, we keep any vertical line segments as possible door posts.
Then, we separate any lines in the top half of the image into three bins based
on whether the line segment has a positive slope, negative slope, or a slope
close to zero, respectively.
An example result is shown in \figref{fig:edges}.

Given the detected edges and the projected wall regions, we can search for
possible doors in the environment.
We iterate over each wall region to find doors in that region.
First, any line segments, of any orientation, that lie outside of the horizontal
range defined by the wall region are removed from consideration.
Then, we iterate over pairs of vertical lines to generate door proposals.
However, we only consider lines that are within a certain distance range of each
other, approximately equivalent to the width of doors.
We approximately localize the vertical lines in 3D space by projecting the raw
LiDAR data to image coordinates, by \equref{equation:calib}, and then
computing the closest LiDAR point to each vertical line.
That LiDAR point's 3D location is used as the approximate location for the
corresponding vertical line.
Then we cluster the vertical lines based on their 3D locations to create a
reduced set of possible door posts represented as
$(u'_{1}, v'_{1}, u'_{2}, v'_{2})$.
Because $u_{1} = u_{2}$ for vertical lines, we compute the average $u_{1}$
within a cluster to compute $u'_{1}$ and $u'_{2}$.
Then we use $u'_{1}$ to find $v'_{1}$ and $v'_{2}$ by computing the
corresponding points on the top and bottom boundaries of the wall region,
respectively.
This creates a smaller set of lines that extend from the bottom boundary to the
top boundary in the wall region on the image.
Using these 3D locations of the lines in this reduced set, we compute a pairwise
distance between them and only keep pairs whose distance is within the range
$[\textrm{0.5~m}, \textrm{1.25~m}]$.

\subsubsection{Scoring proposals}

We introduce a metric to determine how confident we are that a given pair of
vertical lines corresponds to a door or elevator as some proposals could
correspond to signs, posters, or wall structures.
Akin to \citet{DelPero2012BayesianScenes}, we measure how much each proposal
explains, or covers, the detected line segments in the image by computing the
coverage of the vertical lines and top bar (the segment connecting the top
points of the vertical lines) of the proposal.
Multiple line segments may correspond to the same component of the door
proposal, but could be disjoint and/or overlap due to noise or occlusion.
First, we isolate the detected line segments that could correspond to the top of
the door, by only retaining segments whose orientation matches the orientation
of the top boundary of the wall region as well as lie within the top boundary's
vertical tolerance.
Then, using the same approach as for the vertical lines, we compute the
approximate 3D location for each endpoint of each of the remaining line
segments, and only consider line segments whose endpoints are within the
horizontal range: 0.15~m left of the left vertical line and 0.15~m right of the
right vertical line.

With these remaining segments, we can compute what fraction, $c_{\textit{top}}$,
they cover of the top horizontal area between the two vertical line segments.
We take the same approach to the vertical line segments themselves.
For each of the two vertical line segments, we find all of the original line
segments that are within 0.2~m and compute how much they cover those line
segments vertically, resulting in $c_{\textit{left}}$ and $c_{\textit{right}}$.
These three values are aggregated to compute a score for each door as indicated
in \equref{equation:score}.

\begin{equation}
    \label{equation:score}
    \textit{score} =
    \frac{c_{\textit{left}} + c_{\textit{right}} + c_{\textit{top}}}{3}
\end{equation}
Rather than have binary decisions about whether or not a proposal is a door, a
confidence score can allow an online system to be adjusted to a desired
false-positive rate.
In our system, we use a confidence score threshold of 0.75, only considering
detections with a score greater than or equal to this.

\subsubsection{Localizing detections}

Using the 3D locations of the door posts, we describe the location of the door
as the tuple $d =
(x_{\textit{min}}, y_{\textit{min}}, x_{\textit{max}}, y_{\textit{max}})$ where
min and max correspond to the closer and further door posts respectively.
This allows for driving to both sides of the door as the door tag might be on
either side.
Because this process is performed in an online fashion on continuous image
frames received from the camera, the same door can be detected multiple times
at different time steps.
Therefore, detections from different time steps are hierarchically clustered by
the Euclidean distances between their center locations.
For accurate clustering, we rely on additional information from the LiDAR data
to accurately cluster door detections.
For each door bounding box, we find all the 3D points whose corresponding pixel
coordinates are within its bounds.
We then take the median of these 3D points as the door detection's center
location.
This allows for computing accurate door coordinates and filtering out false
positives for driving-goal generation.

\subsubsection{Driving-Goal Generation}

\begin{figure}
  \centering
  \begin{tabular}{@{}cc@{}}
    \begin{subfigure}[t]{0.45\linewidth}
      \includegraphics[width=\textwidth]{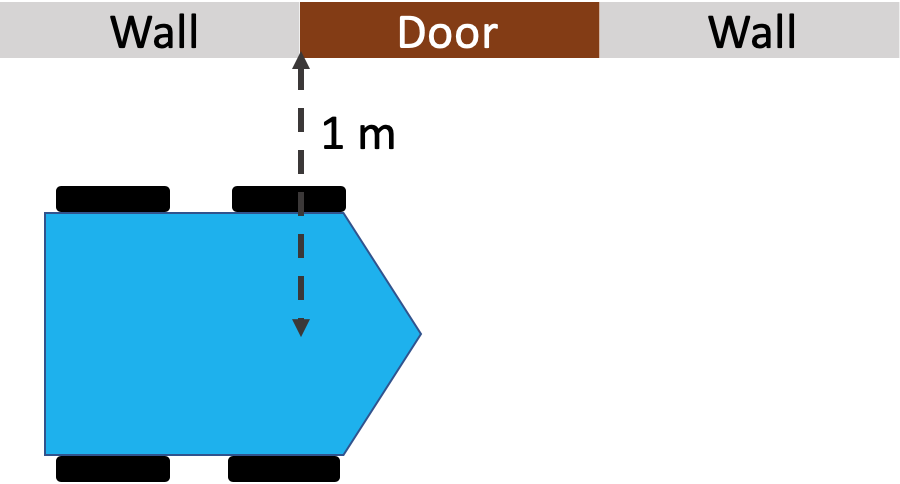}
      \caption{The close side of a door.}
    \end{subfigure}&
    \begin{subfigure}[t]{0.45\linewidth}
      \includegraphics[width=\textwidth]{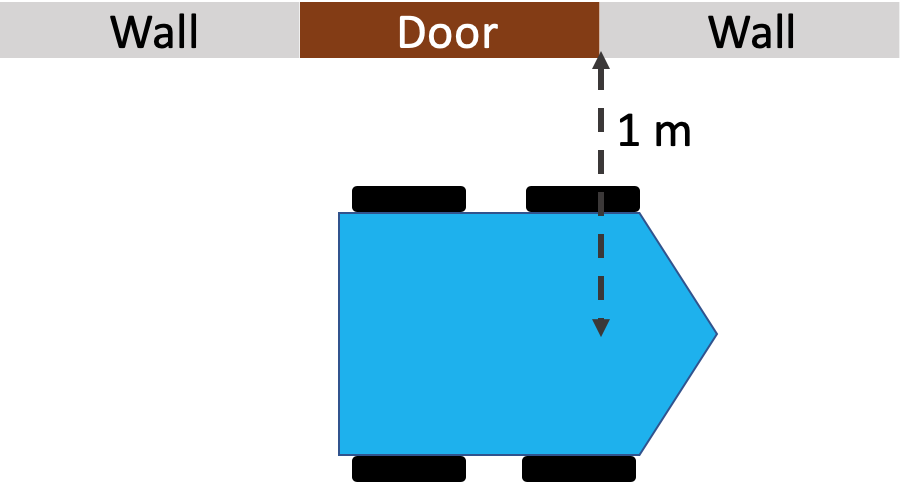}
      \caption{The far side of a door.}
    \end{subfigure}
  \end{tabular}
  \caption{Driving-goal positions for a door.}
  \label{fig:door_positions}
\end{figure}

For each of the resulting clusters, we compute the average locations of the
door posts to create the final driving goals for that door.
Only clusters whose size is $>$3 are considered.
These clusters are classified as being on the \emph{right side} of the hallway
or the \emph{left side} based on a comparison between the robot's trajectory
down the hallway and the door locations (relative to the beginning of the
hallway).
Within these classifications, the clusters are sorted by increasing distance
relative to the beginning of the hallway.
This allows for driving to a specific door.
For example, driving to the third door on the left can be achieved by
retrieving the coordinates of the door whose index is two, by zero-indexing,
among doors classified as being on the left side of the hallway.
The robot uses the coordinates of the door posts to create two driving
goals so that it can position itself appropriately to read the door tag.
\figref{fig:door_positions} illustrates the two desired positions.
The door's angle $\theta = \tan^{-1}\frac{y_{\textit{max}} -
y_{\textit{min}}}{x_{\textit{max}} - x_{\textit{min}}}$ is computed and used
to determine a driving goal that is 1~m perpendicular to the door and with a
target orientation orientation matching that of the door.
Upon arrival at this driving goal, the robot pans its camera in the direction of
the door to read its door tag with Google's Optical Character Recognition API\@.

\subsubsection{Common-Sense Navigation}

Generally, the robot will drive straight down the center of the hallway, using
the \texttt{forward} command, and only drive up to a door once it is within
3~m.
This behavior results in a good view of doors up ahead.
If the index and/or classification of the desired door is unknown, common-sense
knowledge about room labeling in an office environment is leveraged to guess
what they are, based on the first door tag that is read.
This can enable potentially more efficient goal-finding than exhaustive search
alone.
This common-sense knowledge is based on assumptions about doors being labeled
in increasing or decreasing order, both numerically and alphabetically (in the
case of a letter suffix), and even and odd doors being on different sides
of the hallway.
Given the door tag of the first door the robot inspects, it will use parity
to determine what side the goal door is on.
If the goal door is on the same side, it will compute the expected index
of the goal door and drive down the hallway until it detects that door.
Otherwise, it will start with the first door on the other side of the hallway.
For example, if the desired door is room 335 and the first door tag read is
331, then room 335 will be two doors down on the same side of the hallway.
Alternatively, if the first door tag is 330, only doors on the other side
of the hallway will be inspected.

A number of mechanisms are incorporated to allow the robot to find the goal
door in spite of any of these assumptions being invalidated.
Firstly, at any given point, all the door tags read so far are used to determine
whether or not the goal door was missed.
This is done using the range and trend (whether they are increasing or
decreasing) of the door values to check whether the robot should have come
across the door in its path.
If the door has not been missed, the robot will continue to inspect doors on the
current side of the hallway until it reaches the end of the hallway.
In both of these cases, the robot returns to the start of the hallway and begins
an exhaustive search of all doors in the hallway to find the goal door.
If the robot still has not found the goal door after this exhaustive search,
it returns to the \wander\ state.

\section{System Evaluation}

Our system design was developed and validated in three buildings (EE, MSEE, and
PHYS)\@.
In order to evaluate the generalizability and robustness of our system's
performance, we performed 52 trials in 4 distinct floors in each of three new
buildings (HAMP, KNOY, and ME) it had not been deployed in before.
With three exceptions described below, we froze our software after development
before the evaluation trials.
For these trials, we recruited 13 volunteers to provide directions to the
robot.\footnote{Our original intent was to not have any volunteers but instead
  to have the \wander\ state find people naturally occurring in the
  environment to model real visitors soliciting help in finding their way from
  locals.
  COVID-19 prevented us from doing this.}
These volunteers were untrained users who had never interacted with the robot
before nor were aware of what it understood or was capable of.
Each volunteer assisted with 1--6 trials.

\subsection{Experimental Setup}

For each trial, the robot was placed on a floor of one of the three new
buildings and given a random door number as the goal.
The volunteer was instructed to stand in a location where the robot would
be allowed to \wander\ for some time before seeing them.
If the robot re-entered the \wander\ state at any point in the trial,
the volunteer was instructed to relocate to a position that the robot would
come across after being allowed to wander for some time.
A trial was considered a complete success if the robot reached the correct
goal door and read its door tag.
A trial was concluded and subsequently declared a failure if the robot did
not reach the goal after several attempts.
In some trials, some minor manual intervention was used to partially rotate
the robot or prevent it from crashing into a wall to allow the trial to
continue.\footnote{An online appendix at
  \url{https://github.com/qobi/amazing-race} contains the floor plan of each
  building and describes each trial in detail, including the building, floor,
  volunteer, goal, a transcript of the dialog, the plan extracted, the map
  constructed with the route taken, indication of success or failure, and a
  description of the failure reason upon failure.
  The appendix also includes three videos depicting three complete trials,
  one in each building.}

\begin{table*}
  \centering
  \caption{Results.}
  \label{tab:results}
  \begin{tabular}{@{}c@{}c@{}}
    \begin{subtable}[t]{0.35\linewidth}
      \centering
      \caption{Trial results.}
      \label{tab:trialresults}
      \begin{tabular}{llll}
        \hline
        Building & Successes & Total & Success rate \\
        \hline
        HAMP	&  11  &  17 & 64.7\% \\
        KNOY 	&  13  &  16 & 81.3\% \\
        ME 	&  16  &  19 & 84.2\% \\
        \hline
        all   &  40  &  52 & 76.9\% \\
        \hline
      \end{tabular}
    \end{subtable}&
    \begin{subtable}[t]{0.65\linewidth}
      \centering
      \caption{Behavior success rate.}
      \label{tab:behaviorrate}
      \begin{tabular}{llll}
        \hline
        Behavior&Total Instances&Average instances per trial&Success rate\\
        \hline
        \wander                 &  390        & 7.5 & 0.99 \\
        \approachperson         &  385        & 7.4 & 0.47 \\
        \holdconversation 	&  183        & 3.5 & 0.86 \\
        \followdirections 	&  182        & 3.5 & 0.68 \\
        \navigatedoor       	&  \digit99   & 1.9 & 0.40 \\
        \hline
      \end{tabular}
    \end{subtable}
  \end{tabular}
\end{table*}

\subsection{Trial Results}

The number of successful trials is shown in \tabref{tab:trialresults} on a
per-building basis.
Overall, the system was able to successfully reach the given goal in 76.9\% of
the trials.
Of the 12 floor plans tested, the robot was able to successfully find the goal
at least twice on every floor and usually three or more times (8/12 floors).
This result highlights the ability of our system to generalize to new floor
plans.

One of the characteristics of our system is its ability to recover from
individual behavior failures.
Of the 52 trials, only 3 ($\approx$5.7\%) succeeded without any behavior
failures.
When failures do occur, either from shortcomings of the individual behaviors or
incorrect instructions from volunteers, the robot is able to recover and make
multiple attempts at its task.
\tabref{tab:behaviorrate} shows the success rate of each behavior.
Despite there being multiple instances of each behavior due to failure, the
robot is ultimately able to reach the goal door with the subsequent recovery
from the individual behavior failures.
This ability effectively increased the success rate by a factor of $13.3$.

\ignore{
\begin{table}
  \centering
  \caption{Failure Categorization}
  \label{tab:failurecategories}
  \begin{tabular}{llll}
    \hline
    Category 		&  Reason			& Count \\
    \hline
    Bugs 		& door nav reset		& 5 \\
    Navigation 		& 45 degree hallway 		& 1 \\
    Navigation 		& narrow hallway detection	& 1 \\
    Navigation 		& stuck in narrow hall 		& 1 \\
    Vision 		& failed to detect door		& 1 \\
    Vision 		& poster advertisement		& 1 \\
    Vision 		& adjacent door tag 		& 1 \\
    Localization	& failed to localize door	& 1 \\
    \hline
  \end{tabular}
\end{table}}

\subsection{Observations and Improvements}

As the trials were conducted, many observations were made about each behavior's
interaction with the complex environments in which the robot was evaluated.
Although the system recovers from most individual behavior failures, reducing
the number of these would lead to faster and more efficient goal finding.
In this section, we discuss these observations and how they can potentially
be used as points of improvement in future work.

\subsubsection{\approachperson}

We observed several challenges associated with successfully and safely
navigating to a person to ask them for directions.
YOLOv3 cannot distinguish between people and pictures of people on the walls.
In some trials, this led to the robot repeatedly driving up to walls containing
pictures of people and attempting to initiate \holdconversation.
Potential solutions include using body size as a prior to determine whether
the detected person has the correct size for the detected distance or
incorporating body-pose estimation to help make the distinction.
Also, constraints can be applied to the locations of person detections, such as
having to touch the ground, to eliminate detections of people on posters and
signs from consideration.

Another point of difficulty was detecting and localizing people that are far
away.
People that are small in the camera's field of view are more difficult to
detect.
Even if they are detected, the bounding boxes are small, thus having fewer
corresponding 3D LiDAR points, making it difficult to localize, track, and
classify person tracks accurately.
Similarly, consistent, accurate localization of people was also difficult
when they were occluded by objects such as drinking fountains or trash cans,
or when they were positioned close to a wall.
These objects would often overlap with the person detections, which would
affect the localization of the person themselves as some 3D LiDAR points
overlapping the box would actually correspond to the object or wall instead of
the person.
These issues created tracks that would rapidly switch between being approachable
and not approachable, which caused the system to correspondingly switch between
\wander\ and \approachperson\ rapidly.
This explains the high number of instances of \wander\ and \approachperson\ on a
per-trial basis as well as the low success rate of \approachperson.
The robot would eventually reach the person, but would sometimes do so in an
inefficient manner.

\subsubsection{\holdconversation}

The most common observation made when volunteers were providing instructions to
the robot was their use of terminology and hand gestures that the robot did not
understand.
Some volunteers tried to describe directions using gestures, distances, angles,
and objects.
Future work includes developing a more generalizable and robust conversation
engine that is capable of handling the diversity possible in a set of
navigation instructions.
Additionally, the plan structure could also be expanded to include distances,
angles, objects, landmarks, text, arrows, and signs.
Many of those have multiple modalities that could also be explored;
\eg\ distances could be measured in feet, meters, steps, or time, \ie\ `walk
that way for about 2 minutes.'

Aside from this, there was one trial where Google's Speech-to-Text API
returned a non-ASCII character (\degree), which was not supported by our
message passing, and caused our \holdconversation\ behavior to die.
We manually restarted the \holdconversation\ behavior and the trial continued to
success.

\subsubsection{\followdirections}

Many of the building floors featured complexities not seen during development,
such as open spaces and variation in both hallway width and turn angles.
In particular, one building had hallways that branched off at 45\degree\ angles.
This caused some confusion both in processing and executing the navigation
instructions provided by the person.
Some people would indicate that the robot should drive `straight' when it got to
those intersections while others indicated the robot should `turn.'
When told to drive `straight,' \followdirections\ would detect that it could no
longer drive \emph{\textbf{forward}} after reaching the end of the hallway
where the hallway would veer off at 45\degree\ and thus indicate failure.
When told to turn `left,' \followdirections\ would not detect the \textsf{left}
turn and also indicate failure.
An area of future research includes developing adaptive methods for determining
novel intersection types in new indoor environments.

In a different building, the robot could not detect a particular narrow hallway
opening ($\approx$1.5~m wide when most hallways were 2--4~m wide) due to a
hyperparameter corresponding to the hallway width.
We adjusted it accordingly after the first 4 trials, and left it unchanged for
the remainder of the trials.
Dynamically detecting characteristics like this in novel environments is an
area of future research.

Finally, as the robot is driving around, it is building a quantitative map in
real-time and relies on map updates to determine where there are
registered intersections.
When the conjoining hallways are narrow, the robot sometimes does not have
enough data from its LiDAR to have built a reliable-enough quantitative map to
detect the intersection correctly.
Instead of detecting the intersection, the robot drives past it unaware.
Planned future work includes using a neural network to improve the detection of
intersections and execution of the \followdirections\ behavior in spite of an
incomplete map.

\begin{table}
  \centering
  \caption{System Trials.}
  \label{tab:system-trials}
    \begin{tabular}{llll}
      \hline
      group      & live trials   &   \# of trials & success rate \\
      \hline
      ours       & yes           &             52 & \digit76.9\%  \\
      Birmingham & yes           &             46 & \digit67.4\%  \\
      Munich     & yes           &        \digit1 &      100.0\%  \\
      UMBC       & no            &             14 & \digit71.4\%  \\
      Cornell    & yes           &             46 & \digit76.1\%  \\
      \hline
    \end{tabular}
\end{table}

\begin{table*}
  \centering
  \caption{Comparison with Related Work.}
  \label{tab:comparison}
  \begin{tabular}{@{}c@{}}
    \begin{subtable}{\linewidth}
      \caption{Test environment.}
      \label{tab:test-env}
      \resizebox{\textwidth}{!}{\begin{tabular}{lllll}
          \hline
          group      &   \# of test environments & test environment                                & train/test environments are different   & map provided   \\
          \hline
          ours       &               12 & floor of building                               & yes                                 & no             \\
          Birmingham &                \digit1 & 11~m$^2$ with hallway, 2 offices, and 1 conference room & unspecified                         & no             \\
          Munich     &                \digit1 & downtown Munich                                 & unspecified                         & no             \\
          UMBC       &                \digit1 & floor of building                               & yes                                 & yes            \\
          Cornell    &                \digit1 & 1~km$^2$ outdoor facility with 12 buildings       & unspecified                         & no             \\
          \hline
      \end{tabular}}
    \end{subtable}\\
    \\
    \begin{subtable}{\linewidth}
      \caption{People interaction and direction-giving.}
      \label{tab:people-int}
      \resizebox{\textwidth}{!}{\begin{tabular}{llllll}
          \hline
          group      & interacts with live people         & untrained people    & multi-turn conversation & open-ended conversation      & spoken dialogue   \\
          \hline
          ours       & yes                                & yes                 & yes                     & yes                          & yes                          \\
          Birmingham & yes                                & unspecified         & yes                     & yes                          & no                           \\
          Munich     & yes                                & yes                 & yes                     & no                           & no                           \\
          UMBC       & no\footnote{This paper takes, as the input to their trial, a single,
            open-ended text instruction from an untrained user.}                             & yes                 & no                      & no                           & no                           \\
          Cornell    & no\footnote{This paper takes, as the input to their trial, a single,
            open-ended text instruction from a predefined grammar called ``TBS.''\@}                             & no                  & no                      & no                           & no                           \\
          \hline
      \end{tabular}}
    \end{subtable}\\
    \\
    \begin{subtable}{\linewidth}
      \caption{Directions, goals, and recovery.}
      \label{tab:dir-goal-rec}
      \resizebox{\textwidth}{!}{\begin{tabular}{lllll}
          \hline
          group      & follows directions   & can detect when directions are wrong   & robot detects goal independently   & recovers from failure   \\
          \hline
          ours       & yes                  & yes                                    & yes                                & yes                     \\
          Birmingham & no                   & no                                     & yes                                & yes                     \\
          Munich     & yes                  & no                                     & unspecified                        & yes                     \\
          UMBC       & yes                  & yes                                    & unspecified                        & yes                     \\
          Cornell    & yes                  & no                                     & yes                                & no                      \\
          \hline
      \end{tabular}}
    \end{subtable}
  \end{tabular}
\end{table*}

\subsubsection{\navigatedoor}

The \navigatedoor\ behavior exhibited difficulty in finding the goal door in
some situations.
First, many doors were occluded or only partially visible in the robot's
field of view and the door detection method does not support detecting doors in
these scenarios.
A point for future work could be to train a neural-network-based object
detector to robustly detect doors in spite of occlusion.
Additionally, the network could be trained to detect other objects as
well to enable the robot to find something beyond a room.

Another observation is that although the common-sense reasoning led the robot to
efficiently find the goal in some trials, in many cases, it was disrupted by
both missed and false-positive door detections.
In these scenarios, the \navigatedoor\ behavior resorted to exhaustive search to
find the goal door and was typically able to locate it.
However, on a larger scale, exhaustive search would be impractical so a more
robust reasoning and navigation system is an area of future work.
To help remedy missed doors, we adjusted the threshold for the hierarchical
clustering of door detections from 0.25~m to 0.5~m to create fewer but larger
clusters.
This parameter change was done after the first 7 trials and kept for the
remaining trials.

We also found two issues involving using Google's Optical Character Recognition
(OCR) to read door tags.
The first issue involved misreads, which happened twice in our trials.
The robot drove up to the correct door, but read ``goio'' instead of ``g010.''
The same thing occurred with ``g055'' being misread as ``go55.''
A simple string replacement for commonly mistaken characters/digits would solve
this problem (\eg\ o $\Rightarrow$ 0, i $\Rightarrow$ 1, S $\Rightarrow$ 5,
\etc).
The other challenge with OCR was distinguishing \emph{any text read} from text
read from the door tags.
In one failed trial, the robot drove up to a door and read an advertisement
about an event occurring in a particular room.
The room it was referring to was our goal room (which happened to be about 2~m
away, across the hall).
In another failed trial, the robot failed to detect the goal door, drove up to
the subsequent door, but still read the goal door's door tag.
In both of these cases, the robot erroneously claimed success.
Being able to distinguish between a door tag and other signage or text, as
well as being able to assign a single door tag to a single door, would alleviate
this issue.
We counted the two trials with OCR misreads as successes (since the robot did
successfully accomplish all of the behaviors leading up to finding the correct
door) and counted the two trials with the advertisement and the adjacent door
as the mistaken goal as failures.

Lastly, after the first 24 trials, we discovered an issue where the robot was
not clearing its set of detected door locations in between different instances
of the \navigatedoor\ behavior within the same trial.
This was fixed for the remaining trials as the intention was to only reason
about doors in the current hallway as that is where the goal should be located.
However, in future work, retaining and reasoning about all detected doors could
enable the robot to find the goal without directions as well as return to rooms
it has already visited if given instructions to do so.

\section{Related Work}

Several research groups have constructed a system that accomplishes a
comprehensive task similar to ours.
We first compare and contrast our system to these other comprehensive systems.
Then, we compare our individual behaviors to other research that has
focused on similar behavior-specific tasks.

\subsection{Systems}

In \tabref{tab:system-trials} and \tabref{tab:comparison}, we compare our
system to those of the following groups: Birmingham \citep{hanheide2017robot},
whose task is to find an object within a small office environment; Munich
\citep{bauer2009autonomous}, whose task is to navigate the streets of Munich
and reach a downtown plaza; UMBC \citep{matuszek2010following}, whose task is
to follow natural-language directions to an office on a building floor; and
Cornell \citep{oh2015toward}, whose task is to follow a natural-language
directive (in a specified grammar) to an outdoor goal location.

All five groups perform their task in the real world.
Our results, along with Birmingham and Cornell, are based on a large number of
trials, which helps to support the success rate reported.
Our success rate is comparable to the other groups, but, as we will show in the
subsequent tables, we also take on more challenges.

\Tabref{tab:test-env} describes the test environment and how many distinct
environments each group tested on.
We test in 12 distinct environments, which is much larger than any other group.
Having multiple trials in a large number of distinct environments provides
some additional support to the generalizability of our system.
We also stress that our 12 environments are distinct from the environments
we trained on and developed for.
Doing so prevents us from ``training on (or developing for) the test set,''
which carries the risk of overfitting to a particular environment and biasing
the generalizability of a system's performance.
Like most of the other systems, we operate in an unknown environment, without a
map.

\Tabref{tab:people-int} compares how the robot interacts with people and gets
directions.
Unlike UMBC and Cornell, which have a single set of directions provided to the
robot at the beginning of the trial, we have to find people in our environment
that we can ask for directions.
These people are untrained and unfamiliar with what the robot understands or is
capable of.
Along with Birmingham, we allow for open-ended conversation, which increases the
diversity of responses.
Finally, our conversation takes place via spoken dialogue which no other systems
do.
(Munich uses a GUI to get responses, while Birmingham uses text input.)
These differences make our task considerably harder than those of other systems.

Finally, in \tabref{tab:dir-goal-rec}, we point out a few additional ways in
which our task sets us apart from the other systems.
Our robot has to follow directions, rather than simply exploring until the goal
is found.
Exploring may work when the environment is small, but becomes intractable as the
environment increases in size.
When the directions it has are wrong, both our robot and UMBC's are capable of
detecting this.
In UMBC's case, they backtrack and recompute the next-most-likely path to the
goal.
In our case, we simply revert to \wander\ and seek out another person to ask for
directions.
Our robot is capable of discovering and detecting the goal independently, while
it is unclear whether all systems are capable of that ability.
Lastly, like most other systems, we are able to recover from individual behavior
failures and continue working towards the goal.

\subsection{Behaviors}

As described earlier, our system combines a number of behaviors, each of which
solve one of the tasks presented by The Amazing Race\TM\ challenge.
The first is autonomous navigation in an indoor environment.
\citet{barrett2018driving} employ a learning mechanism to acquire the semantics
for direction keywords such as `left of,' `right of,' and `behind' and then use
these semantics for planning and describing robot paths.
Because our focus was to design a complete system for navigation in an unknown
environment, we predefined a similar set of directions for the
\wander\ and \followdirections\ state and used a novel method for determining
whether they are available.

The next crucial task is to find a person and receive and interpret directions
from them.
The Jackrabbot lab at Stanford has developed a robot that complies with social
conventions such as understanding \citep{robicquet2016learning} and predicting
\citep{alahi2016social} human trajectory in crowded scenes.
When looking for and approaching people, our robot also complies with expected
conventions such as determining which people are approachable, introducing
itself as it approaches, and not invading personal space when having a
conversation.

\citet{bauer2009autonomous, bauer2009heuristic, bauer2009information} define a
set of heuristic rules and a complex finite-state machine to obtain specific
pieces of information from a person through a touch screen interface on their
robot.
Our system, however, takes full spoken natural-language utterances as input and
extracts instructions from them.
\citet{oh2015toward} and \citet{boularias2015grounding} require a specific
syntactic structure of natural-language commands called Tactical Behavior
Specification to simplify the parsing problem.
Our parsing method does not depend on such a strict grammar because it would
be impractical for interacting with people unfamiliar with the robot.
\citet{kollar2010toward} train a model to extract spatial-description clauses
from natural-language directions to determine the corresponding path in the
environment.
Our method does not rely on training.

A single person's response may have insufficient, vague, or incorrect
information.
\citet{thomason2015learning} present a dialogue system that can compensate for
this issue by generating queries about missing pieces of information,
specifically the action, patient, recipient, or some combination of them.
Our dialogue system does not predefine what or how much information we seek and
instead dynamically infers and generates queries for what information is
missing.

In order to navigate in the environment, given the natural-language directions,
the robot needs a semantic map of the environment.
\citet{hemachandra2015learning} rely on AprilTag fiducials to classify regions
in the environment for autonomous navigation.
However, this prevents practical operation in any unknown environment because
it would have to be labeled with these fiducials.
\citet{sunderhauf2016place} explore new areas and create a semantic map
through the classification of sequences of image frames as places (\eg\ office
or kitchen).
\citet{pangercic2012semantic} create semantic object maps in a kitchen using
a RGBD camera that enables their robot to perform fetch and place tasks.
We use the occupancy grid, door detection, and text recognition to make a
semantic map of an unknown environment.

Many approaches, including ours, rely on edge or line segment detection
as a basis for finding door posts.
\citet{Stoeter2000Real-timeEnvironments} detects door frames by detecting
vertical stripes in the image, but other objects or parts of the background
could be responsible for vertical edges.
More complex approaches look for edges in the image that form an upside-down
U-shape to create door detections
\citep{Kim1998RecognitionFunctionality, Shi2006InvestigatingEnvironments}.
These methods depend on high quality detected edges, which are not necessarily
possible to obtain especially on images captured from a camera on a robot
navigating in areas with varying, possibly poor, lighting conditions.
Our method incorporates noise-tolerance mechanisms that simultaneously allow for
door detection in spite of poor edge detection and measuring the confidence
of those detections.

In the case of \citet{Shi2006InvestigatingEnvironments}, only U-shapes that
intersect the corridor lines are considered doors in order to avoid false
positives caused by signs or posters.
These corridor lines are obtained by finding the vanishing lines that have the
most intersections with the bottoms of vertical lines.
\citet{Olmschenk3DImage} employ a similar method to finding corridor lines by
detecting the vanishing point and vertical lines in an image to determine where
the wall meets the floor.
This approach is impractical for two reasons:
\begin{inparaenum}
\item walls orthogonal to the viewpoint are ignored and
\item patterns in the image, such as floor tiling, can induce strong edge
  detections that can be confused for corridor lines.
\end{inparaenum}
We incorporate 3D information from LiDAR data to reliably and generally
detect wall regions and avoid this issue.

\citet{hanheide2017robot} employ a knowledge hierarchy that enables reasoning
over known pieces of information in order to perform task planning, execution,
and task-failure explanation.
Because our robot has no prior knowledge about the environment, we extract
information such as driving directions and door locations in an online fashion
to execute navigation instructions and plan paths to potential locations of the
goal.

\section{Conclusion and Future Work}
\label{sec:conclusion-future-work}

In this manuscript, we proposed The Amazing Race\TM: Robot Edition task in
which a robot must find any designated location.
We described a system design that solves this task with the
scope of finding a specific room on a single floor of a building.
We demonstrated that our system architecture lets
a robot complete this task by a making a series of logical steps akin to
those a human would make.
Our robot autonomously finds a person in the environment and engages them in
dialogue for directions.
It then follows these directions to reach the hallway in which it then searches
the doors to find the designated goal.

We demonstrated that our system was successful in 76.9\% of the trials that we
conducted.
As discussed, future work entails making improvements to each of the behaviors
in our system to both improve the success rate and increase the scope of
the task.

\appendices

\section*{Acknowledgment}

Any opinions, findings, views, and conclusions or recommendations expressed in
this material are those of the authors and do not necessarily reflect the
views, official policies, or endorsements, either expressed or implied, of the
sponsors.
The U.S. Government is authorized to reproduce and distribute reprints for
Government purposes, notwithstanding any copyright notation herein.

\bibliographystyle{abbrvnat}
\bibliography{tro2020}

\begin{IEEEbiography}
  [{\includegraphics[width=1in,height=1.25in,clip,keepaspectratio]
      {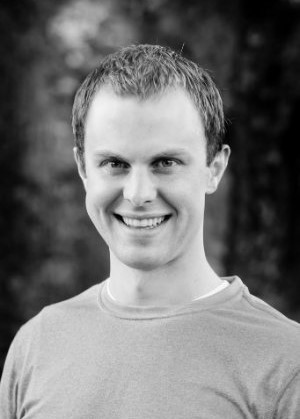}}]{Jared S. Johansen} received a B.S. in Electrical
  Engineering from Brigham Young University (2010) and an M.S. in Electrical
  Engineering and an MBA from the University of Utah (2012).
  He is currently a Ph.D. student in the School of Electrical and Computer
  Engineering at Purdue University.
  His research lies at the intersection of artificial intelligence, machine
  learning, computer vision, natural-language processing, and robotics.
\end{IEEEbiography}

\begin{IEEEbiography}
  [{\includegraphics[width=1in,height=1.25in,clip,keepaspectratio]
      {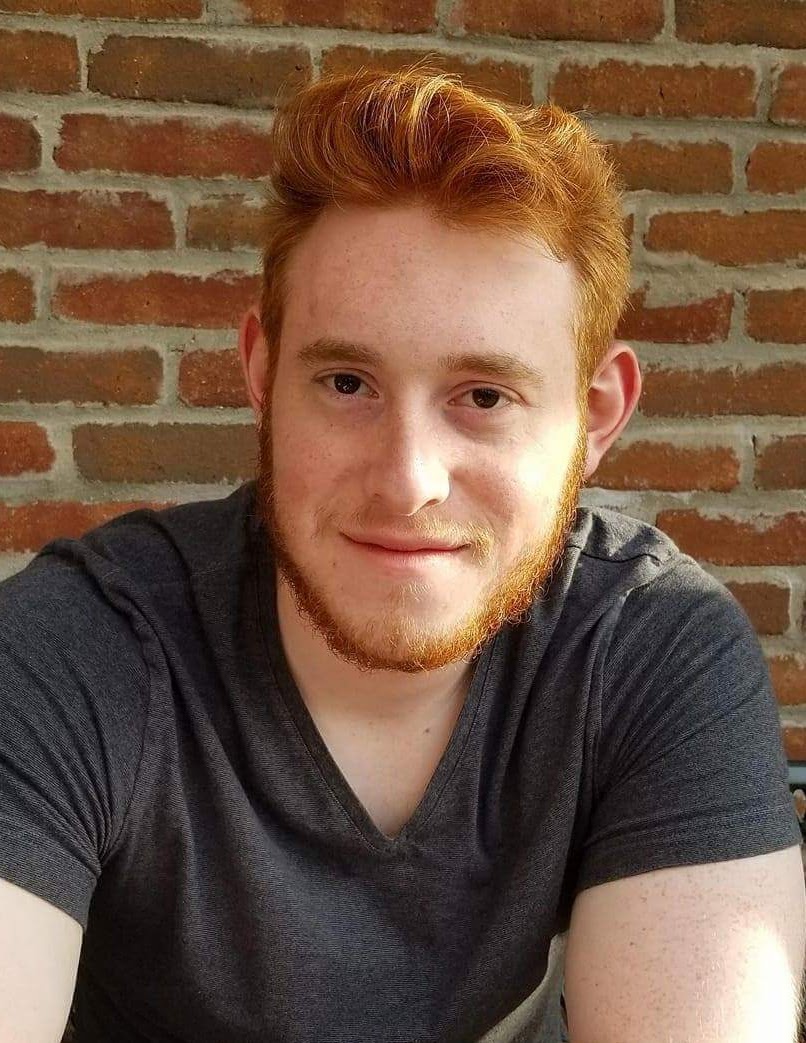}}]{Thomas V. Ilyevsky} received his B.S. in Electrical
  and Computer Engineering from Cornell University in 2016.
  He is currently pursuing a Ph.D. in Electrical and Computer Engineering at
  Purdue University.
  His research focuses on artificial intelligence, computer vision, and
  natural-language processing in the context of human-computer interaction.
\end{IEEEbiography}

\begin{IEEEbiography}
  [{\includegraphics[width=1in,height=1.25in,clip,keepaspectratio]
      {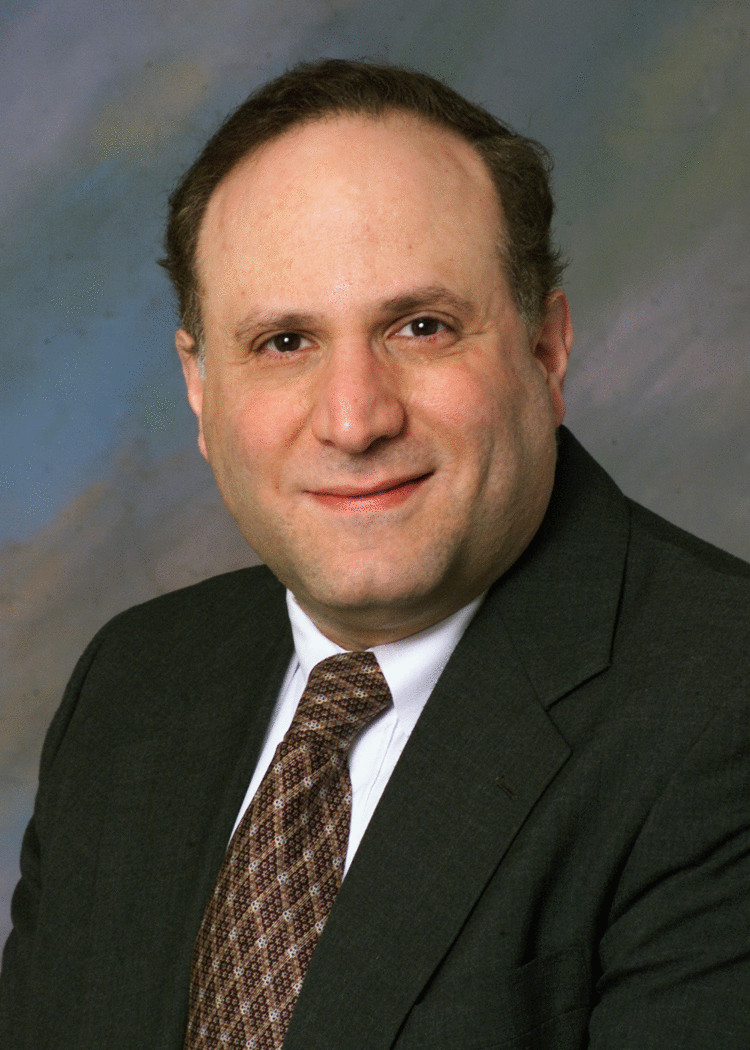}}]{Jeffrey Mark Siskind} received the B.A. degree in
  Computer Science from the Technion, Israel Institute of Technology in 1979,
  the S.M. degree in Computer Science from MIT in 1989, and the Ph.D. degree
  in Computer Science from MIT in 1992.
  He did a postdoctoral fellowship at the University of Pennsylvania
  Institute for Research in Cognitive Science from 1992 to 1993.
  He was an assistant professor at the University of Toronto Department of
  Computer Science from 1993 to 1995, a senior lecturer at the Technion
  Department of Electrical Engineering in 1996, a visiting assistant professor
  at the University of Vermont Department of Computer Science and Electrical
  Engineering from 1996 to 1997, and a research scientist at NEC Research
  Institute, Inc.\ from 1997 to 2001.
  He joined the Purdue University School of Electrical and Computer
  Engineering in 2002 where he is currently a professor.
\end{IEEEbiography}

\end{document}